%% file: main.tex
\documentclass[10pt,twocolumn,letterpaper]{article}

\usepackage{cvpr}              

\input{preamble}

\definecolor{cvprblue}{rgb}{0.21,0.49,0.74}

\usepackage{pifont}
\usepackage{color}
\usepackage{wrapfig}
\usepackage{makecell}
\usepackage{colortbl}
\usepackage{multirow}
\usepackage{fixltx2e}
\usepackage{subcaption}
\usepackage{etoolbox}
\usepackage{multicol}
\usepackage{refcount}
\usepackage{enumitem}
\usepackage[misc]{ifsym}
\usepackage{subcaption}

\definecolor{LightCyan}{rgb}{0.88,1,1}
\definecolor{mygray}{gray}{0.9}
\definecolor{mygray2}{gray}{0.6}

\usepackage[pagebackref,breaklinks,colorlinks,allcolors=cvprblue]
{hyperref}

\def\paperID{13178} 
\def\confName{CVPR}
\def\confYear{2025}

\title{Universal Actions for Enhanced Embodied Foundation Models}

\author{
Jinliang Zheng\footnotemark[1]~~$^{1,2,5}$, Jianxiong Li\footnotemark[1]~~\footnotemark[2]~~$^{1}$, Dongxiu Liu\footnotemark[1]~~$^{4,1}$, {Yinan Zheng}$^{1}$,\\ {Zhihao Wang}$^{3,1}$, {Zhonghong Ou}$^{4}$, {Yu Liu}$^{2}$, {Jingjing Liu}$^{1}$, {Ya-Qin Zhang}$^{1}$, {Xianyuan Zhan}\footnotemark[3]~~$^{1, 5, 6}$\\
$^1$ AIR, Tsinghua University, 
$^2$ Sensetime Research,
$^3$ Peking University\\
$^4$ Beijing University of Posts and Telecommunications,
$^5$ Shanghai AI Lab, $^6$ BAAI\\
\texttt{\{zhengjl23, li-jx21\}@mails.tsinghua.edu.cn}\\
\texttt{zhanxianyuan@mail.tsinghua.edu.cn}\\
}

\begin{document}
\maketitle

\renewcommand{\thefootnote}{\fnsymbol{footnote}}
\footnotetext[1]{Equal contribution}
\footnotetext[2]{Project Lead}
\footnotetext[3]{Corresponding author}

\input{sec/0_abstract}    
\input{sec/1_intro}

\input{sec/2_formatting}
\input{sec/3_finalcopy}
{
    \small
    \bibliographystyle{ieeenat_fullname}
    \bibliography{main}
}
\newpage

\appendix
\section{Training Details}
\label{sec:train_appendix}
\noindent\textbf{Training hyper-parameters} UniAct-0.5B utilizes the pretrained parameters from LLava-One-Vision-0.5B~\cite{li2024llava} to initialize the VLM. For visual feature extraction in heterogeneous decoding heads, we deploy an ImageNet pretrained ResNet18, which is commonly employed in vision-based policy learning~\cite{walke2023bridgedata}. This model was jointly trained during the large-scale pre-training process to enhance perceptual capability in manipulation task scenarios. We adopt resolutions of 
384×384 for the VLM and 
224×224 for the ResNet18, consistent with their original configurations. Image augmentation settings and more hyper-parameters can be found in Table~\ref{tab:image_augmentation} and Table~\ref{tab:pretraing_setting}, respectively.

\begin{table}[ht]
    \centering
    \begin{tabular}{l|c}
        \toprule
        Augmentation & value \\
        \midrule
        \multirow{3}{*}{RandomResize} & ratio=(0.75, 1.3333)  \\
             & scale=(0.5, 1.0) \\ 
             & interpolation=BICUBIC \\
        \midrule
        RandomHorizontalFlip & p=0.5 \\
        \midrule
        \multirow{3}{*}{ColorJitter} & contrast=(0.6, 1.4) \\
         & brightness=(0.6, 1.4) \\
         & saturation=(0.6, 1.4)\\
        \bottomrule
    \end{tabular}
    \caption{Image augmentation settings during training}
    \label{tab:image_augmentation}
\end{table}

\begin{table}[h]
    \centering
    \begin{tabular}{l|c}
        \toprule
        config & value \\
        \midrule
        optimizer & AdamW  \\
        batch size &  1024 \\
        learning rate & $2\times10^{-5}$ \\
        weight decay & 0. \\
        optimizer momentum & $\beta_1,\beta_2$=0.9,0.95  \\
        iters & 500K \\
        model precision & BFloat16 \\
        \bottomrule
    \end{tabular}
    \caption{Training hyper-parameters}
    \label{tab:pretraing_setting}
\end{table}

\noindent\textbf{Data construction}
As illustrated in Table~\ref{tab:data}, we detail the composition of our training data, which includes the number of trajectories, samples, and the control interfaces for the 28 distinct embodiments. Following previous works~\cite{kim2024openvla, team2024octo}, we assign different sampling rates to each dataset during training to ensure a balanced mix of embodiments, tasks, and scenes. These sampling rates are specified in Table~\ref{tab:data}. It is important to note that many of the datasets may contain distinctly action spaces such as EEF position and Joint position. In addition, images from different datasets may contain multiple view points. By default, we use only the first third-person perspective following~\cite{kim2024openvla, team2024octo} to maintain consistency.

\begin{table*}[h]
    \centering
    \begin{tabular}{l|c|c|c|c}
        \toprule
        Dataset & Trajectory & Samples & Sample rate(\%) & Control Interface \\
        \midrule
        Utaustin Mutex & 1500 & 361871 & 1.0 & EEF Position \\
        Berkeley Cable Routing & 1557 & 38789 & 0.2 & EEF velocity\\
        NYU Franka Play & 454 & 44412 & 1.0 & EEF velocity \\
        Kuka  & 557893 & 7130157 & 12.7 & EEF Position\\
        Austin Sailor & 240 & 352110 & 2.2 & EEF velocity \\
        Fmb & 8611 & 1136907 & 1.0 & EEF velocity \\
        Berkeley Autolab & 1000 & 95310 & 1.2 & EEF Position \\
        Viola & 150 & 75614 & 0.9 & EEF Position \\
        Dobbe & 5208 & 1139874 & 1.4 & EEF Position \\
        Iamlab CMU & 631 & 146241 & 0.9 & EEF Position \\
        Austin Buds & 50 & 34110 & 0.2 & EEF Position \\
        Language Table & 441911 & 6602077 & 4.4 & EEF Position \\
        Stanford Hydra & 569 & 357137 & 4.4 & EEF Position \\
        Robo Set & 18246 & 1419838 & 5.0 & Joint position \\
        Austin Sirius & 559 & 279724 & 1.7 & EEF velocity \\
        Dlr Edan & 104 & 8824 & 0.1 & EEF Position \\
        Fractal & 86599 & 3607028 & 12.7 & EEF Position \\
        TOTO & 1003 & 324669 & 2.0 & Joint position \\
        Berkeley Fanuc & 415 & 58660 & 0.7 & EEF Position \\
        CMU Stretch & 135 & 25012 & 0.2 & EEF Position \\
        Roboturk & 1934 & 186910 & 2.3 & EEF Position \\
        Jaco Play & 976 & 66094 & 0.4 & EEF Position \\
        Taco Play & 3603 & 230966 & 3.0 & EEF Position \\
        BC-Z & 42811 & 5957097 & 7.5 & EEF Position \\
        Droid & 92115 & 27043929 & 10.0 & EEF Position \\
        Furniture Bench & 5100 & 3905717 & 2.4 & EEF velocity \\
        Bridge & 28933 & 899685 & 13.3 & EEF Position \\
        Libero & 6500 & 1007204 & 5.0 & EEF Position \\
        \bottomrule
    \end{tabular}
    \caption{Details of the training data composition, including the number of trajectories, number of samples, and control interfaces of the 28 different data sources. Following~\cite{kim2024openvla}, we set different sampling probabilities for different data, which we also report in this table.}
    \label{tab:data}
\end{table*}

\noindent{\textbf{Categorical Reparameterization}}.
Except for the Gumbel-Softmax utilized for training UniAct, we also explored another commonly used reparameterization technique: the Straight-Through Estimator (STE)~\cite{van2017neural}. However, empirical findings indicate that using STE can lead to severe collapses in the universal action codebook. An intuitive explanation for this is that each universal action requires numerous optimization steps to learn the highly abstracted behaviors. Since STE involves hard sampling, it results in only one universal action being selected and optimized for each training sample. This lack of gradient distribution among all potential actions can stifle the learning process and reduce the diversity of learned actions.

\section{Evaluation Setups}
\label{sec:eval_appendix}
In this section, we delve deeper into the evaluation experiments for the three robotic embodiments used in our study: WidowX Robot, Franka Robot, and AIRBOT. We provide detailed scores for each embodiment to offer a clearer, more intuitive comparison of their performance.

\subsection{WidowX Robot in Real World}

\begin{figure*}[t]
    \centering
    \includegraphics[width=0.99\linewidth]{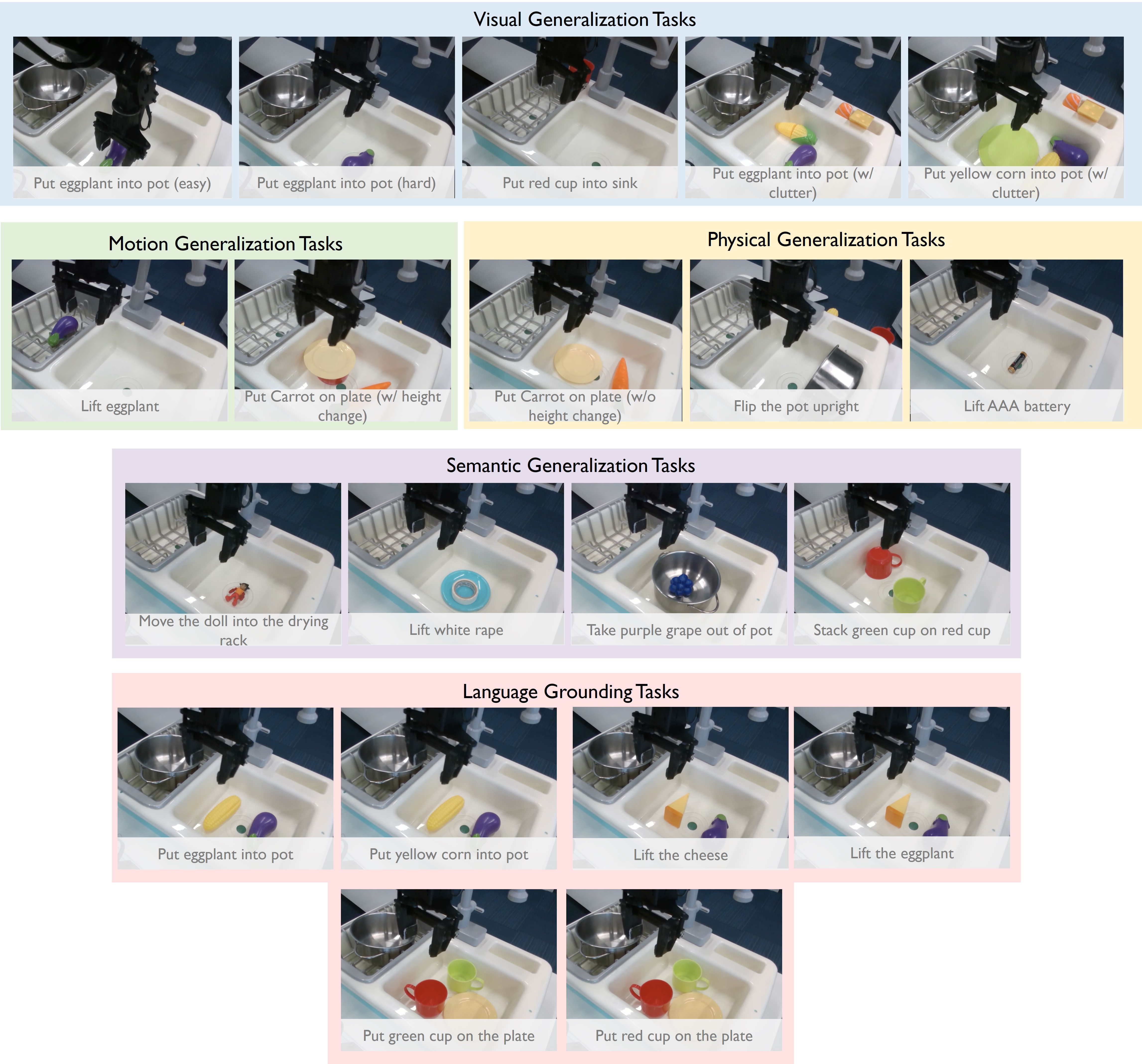}
    \caption{Illustration for WidowX evaluation tasks.}
    \label{fig:widowx_tasks}
\end{figure*}

The experiments on WidowX aim to assess the models' generalization capabilities across five distinct dimensions, as illustrated in Figure~\ref{fig:widowx_tasks}. Instead of merely reporting task success rates, we calculate scores based on the progress made towards completing the task for some complex tasks. This scoring method can more intuitively reflect the model’s understanding and generalization ability~\cite{kim2024openvla}. Detailed scores are available in Table~\ref{tab:widowX_detailed_comparison}. In the subsequent sections, we will provide a comprehensive description of the task settings. 

\noindent{\textbf{Baseline models}}: Given that Octo, CrossFormer, and OpenVLA were trained using same data recipes (e.g., OXE) as UniAct, we directly evaluated their performance on the WidowX robot without further fine-tuning, consistent with their original implementations. For LAPA, we employed the OXE-pretrained model and adhered to the official protocol for fine-tuning with Bridge Data. This fine-tuning process involved 2,000 gradient steps with a batch size of 64. An intriguing observation is that, despite CrossFormer's utilization of more diverse pretraining data, its performance was inferior to that of its codebase model, Octo. We hypothesize that this discrepancy stems from a potentially suboptimal training pipeline, wherein all single-arm robots, regardless of their control interfaces, were treated as a homogeneous embodiment. This approach may have induced negative transfer, resulting in performance degradation. In contrast, while UniAct also leverages diverse and heterogeneous data, similar to CrossFormer, its well-defined universal action space effectively mitigates negative transfer.

\noindent{\textbf{Visual Generalization}}:
This dimension evaluates the model’s ability to adapt to different visual environments characterized by variations in lighting, background, and object textures. Following the setups outlined in~\cite{kim2024openvla}, we design three specific tasks, detailed in Figure~\ref{fig:widowx_tasks}. Each task is set against a distinct background environment, features different lighting brightness levels, and involves target objects in two colors: red and green. Additionally, we introduce various visual distractors unrelated to the task to assess the model's ability to generalize visually and maintain robustness against such distractions. To ensure a fair comparison, all variables related to the model's testing conditions are kept consistent across all models. All tasks in this suite are assigned binary score: 0 for failure and 1 for success.

\noindent{\textbf{Motion Generalization}}:
Tasks in this dimension aim to evaluate the robot's capability to perform appropriate motions while recognizing the target object's position and orientation, which may not have been encountered during training. Specifically, we have designed two tasks: \textit{Lift eggplant} and \textit{Put carrot on plate}, as detailed in Figure~\ref{fig:widowx_tasks}. The target objects are placed in several predetermined positions and oriented in pre-designed directions. This helps in assessing the robot's adaptability to changes in physical task parameters.  All tasks in this suite are assigned binary score: 0 for failure and 1 for success.


\noindent{\textbf{Physical Generalization}}:
In this dimension, we examine the model's capability to manage physical variations in objects, such as size, weight, and material properties. The tasks are crafted to evaluate the robot's adaptability to these fluctuations, which significantly influence manipulation strategies. Detailed task setups are listed in Figure~\ref{fig:widowx_tasks}. The objects involved in these tasks, such as carrots and AAA batteries, often have irregular shapes or are placed in unusual positions (e.g., an overturned pot). Successfully handling and completing tasks with these objects requires a generalized policy that can accurately recognize the physical attributes of the target objects and execute appropriate strategies. All tasks in this suite are assigned binary score: 0 for failure and 1 for success.


\noindent{\textbf{Semantic Generalization}}:
In this dimension, we assess the model's ability to understand and generalize across various semantic contexts. Specifically, the tasks especially the target objects included in this dimension have never been encountered during the training procedures for both UniAct and the baseline models. This approach tests the model’s capability to interpret and adapt to new instructions and environments, emphasizing its flexibility and learning efficiency. The purple grape is highly smooth, so we assign scores as follows: 0.25 for touching the grape, 0.5 for grasping it, 0.75 for moving it toward the pot, and 1 for successfully completing a pick-and-place. Also, for the stack green cup on red cup task, we assign scores as follows: 0.25 for touching the green cup, 0.5 for grasping it, 0.75 for moving it toward the red cup, and 1 for successfully stacking. Other tasks in this suite are assigned binary scores.



\noindent{\textbf{Language Grounding}}:
This dimension evaluates the model's ability to comprehend and execute commands that are grounded in natural language. The focus is on assessing the model’s proficiency in following novel instructions to manipulate specific objects as described verbally. While there are similarities with Semantic Generalization in terms of handling unseen scenarios, Language Grounding task distinctly tests the model's capacity to accurately understand and act on language-based directives within potentially misleading environments.
An illustrative example is placing two cups of different colors on a table and instructing the model to manipulate one of them using language. The model must accurately ground the language instruction to the correct object in a complex real-world setting. This tests the model’s ability to connect linguistic descriptions directly with physical actions in dynamic and visually diverse environments. For all tasks in this suite, we assign scores as follows: 0.5 for correct grounding, 1 for a successful task completion.

\begin{table*}[h]
    \centering
    \begin{tabular}{l|c|c|c|c}
        \toprule
        Task Category & Task Name & Octo & OpenVLA & UniAct-0.5B(ours) \\
        \midrule
        \multirow{6}{*}{Visual Generalization} 
            & \textit{put eggplant into pot(easy)} & 3.0 & \textbf{10.0} & \textbf{10.0} \\
             & \textit{put eggplant into pot} & 6.0 & 6.0 & \textbf{10.0} \\ 
             & \textit{put cup from counter into sink} & 0.0 & \textbf{2.0} & 1.0 \\
             & \textit{put eggplant into pot} & 3.0 & \textbf{8.0} & 5.0 \\
             & \textit{put corn on plate} & 0.0 & 7.0 & \textbf{8.0} \\
             \rowcolor[gray]{.9} 
             & Average Score & 2.4 & 6.6 & \textbf{6.8} \\
        \midrule
        \multirow{3}{*}{Motion Generalization} 
            & \textit{lift eggplant} & 4.0& 3.0& \textbf{6.0} \\
             & \textit{put carrot on plate} & 2.0& 4.0 &\textbf{6.0} \\ 
             \rowcolor[gray]{.9} 
             & Average Score & 3.0 & 3.5 & \textbf{6.0} \\
        \midrule
        \multirow{4}{*}{Physical Generalization} 
            & \textit{put carrot on plate} &2.0 &\textbf{9.0} &\textbf{9.0} \\
             & \textit{flip pot upright} &1.0 & 2.0& \textbf{7.0}\\ 
             & \textit{lift AAA battery} &0.0 &\textbf{7.0} & 5,0\\
             \rowcolor[gray]{.9} 
             & Average Score & 1.0 & 6.0 & \textbf{7.0} \\
        \midrule
        \multirow{5}{*}{Sematic Generalization} 
            & \textit{move doll into drying rack} &0.0 & 5.0 & \textbf{6.0} \\
             & \textit{lift white rape} &0.0 & 1.0 & \textbf{2.0}\\ 
             & \textit{take purple grapes out of pot} &4.5 & \textbf{5.0} & \textbf{5.0} \\
            & \textit{stack green cup on red cup} & 3.25 & \textbf{8.25} & 2.5 \\
            \rowcolor[gray]{.9} 
             & Average Score & 1.9 & \textbf{4.8} & 3.9 \\
        \midrule
        \multirow{7}{*}{Language Grounding} 
            & \textit{put eggplant into pot} & 6.0 & 7.0 & 5.0  \\
             & \textit{put yellow corn into pot} & 7.0 & 10.0 & 10.0 \\ 
             & \textit{lift cheese} &1.0 & \textbf{8.0} & 6.0 \\
             & \textit{lift eggplant} & 6.0 & 8.0 & \textbf{9.0} \\
             & \textit{put green cup on plate} &3.0 & \textbf{10.0} & \textbf{6.0} \\
             & \textit{put red cup on plate} & 6.0 & \textbf{10.0} & \textbf{8.0} \\
             \rowcolor[gray]{.9} 
             & Average Score & 4.8 & \textbf{8.8} & 7.3 \\
        \bottomrule
    \end{tabular}
    \caption{Comparison between UniAct-0.5B and baseline models on detailed scores in WidowX evaluations.}
    \label{tab:widowX_detailed_comparison}
\end{table*}

\subsection{Franka Robot in Simulation}

\begin{figure*}[t]
    \centering
    \includegraphics[width=0.3\textwidth,height=0.25\textwidth]{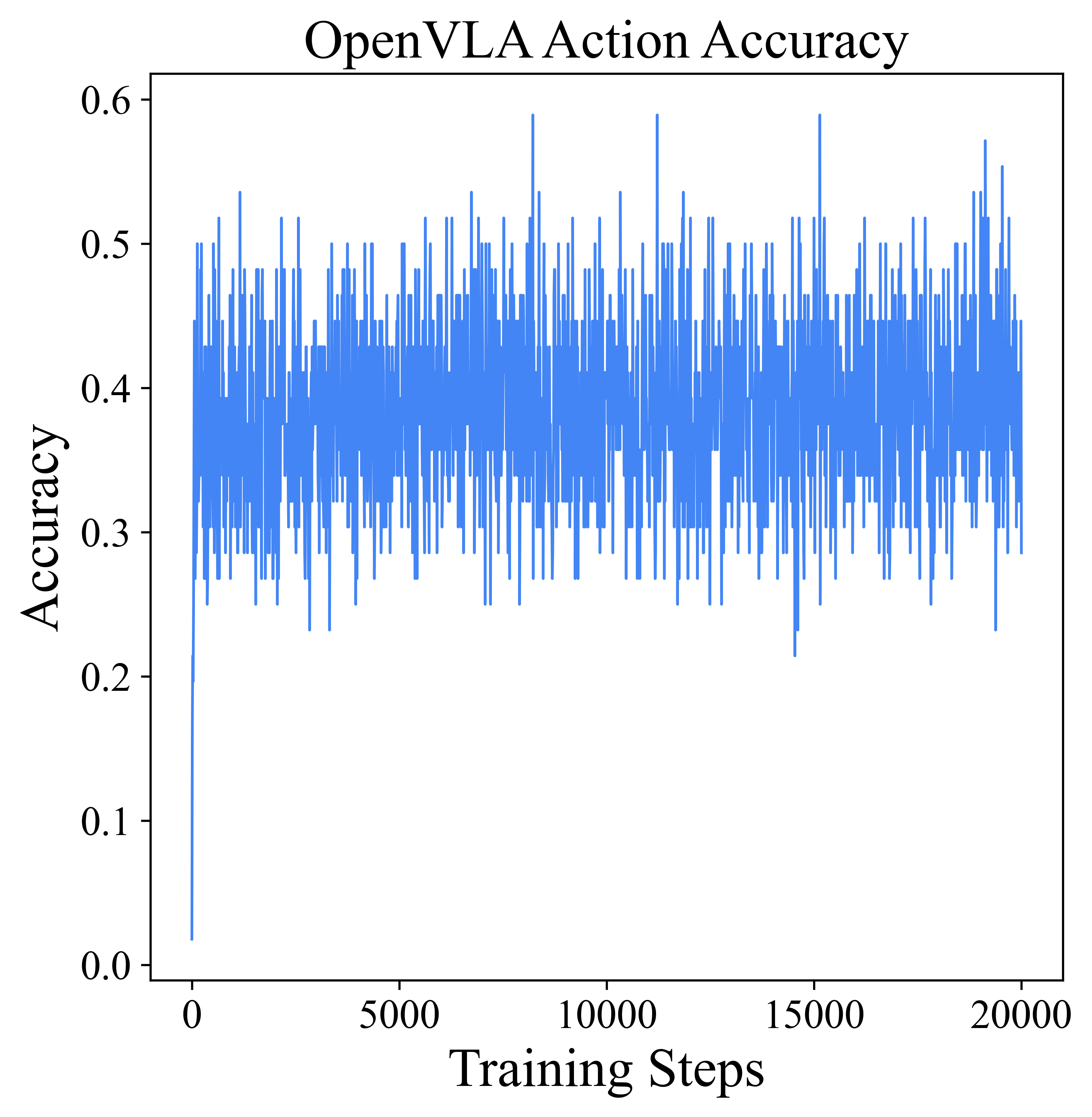}
    \includegraphics[width=0.3\textwidth,height=0.25\textwidth]{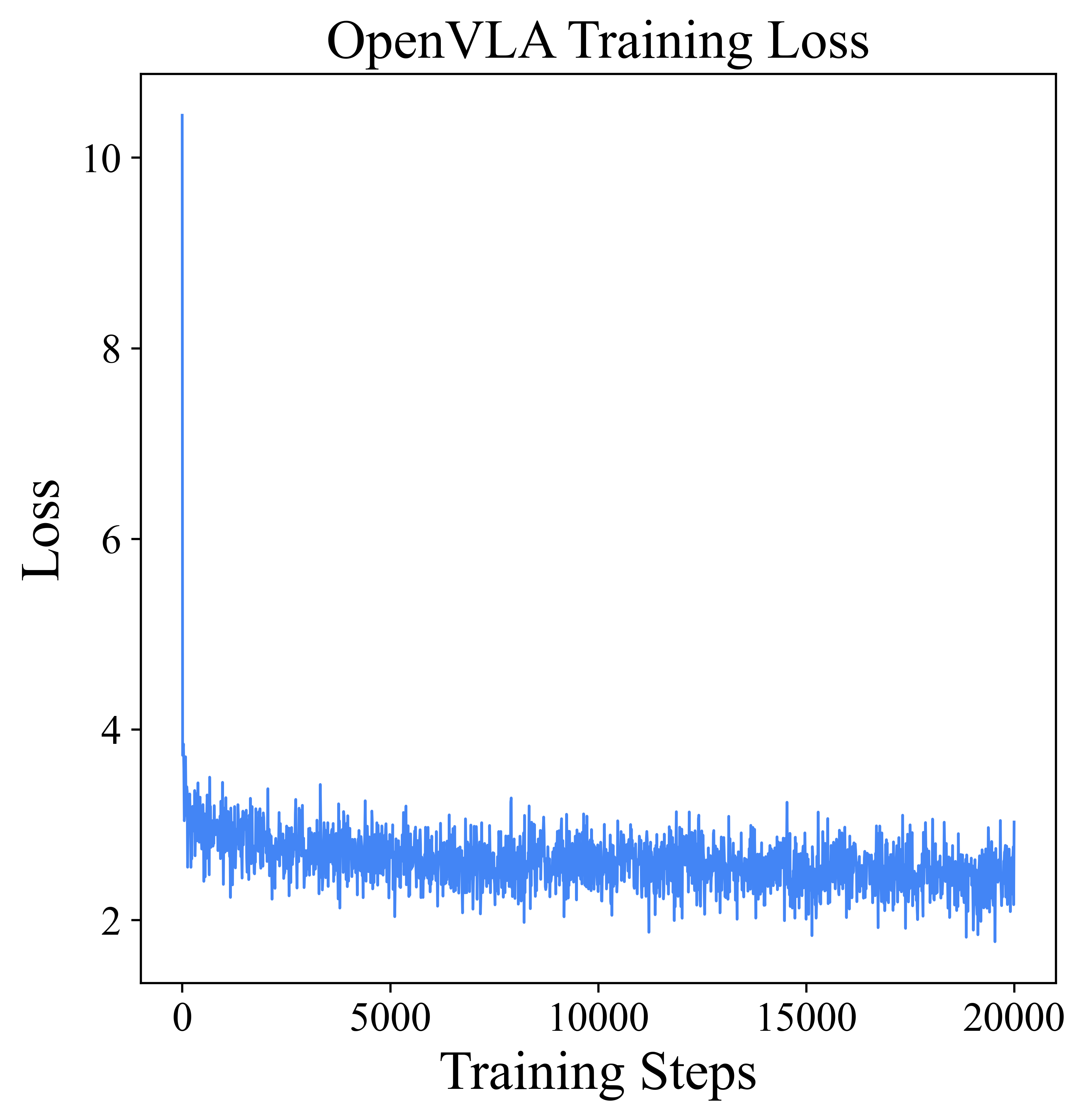}
    \includegraphics[width=0.3\textwidth,height=0.25\textwidth]{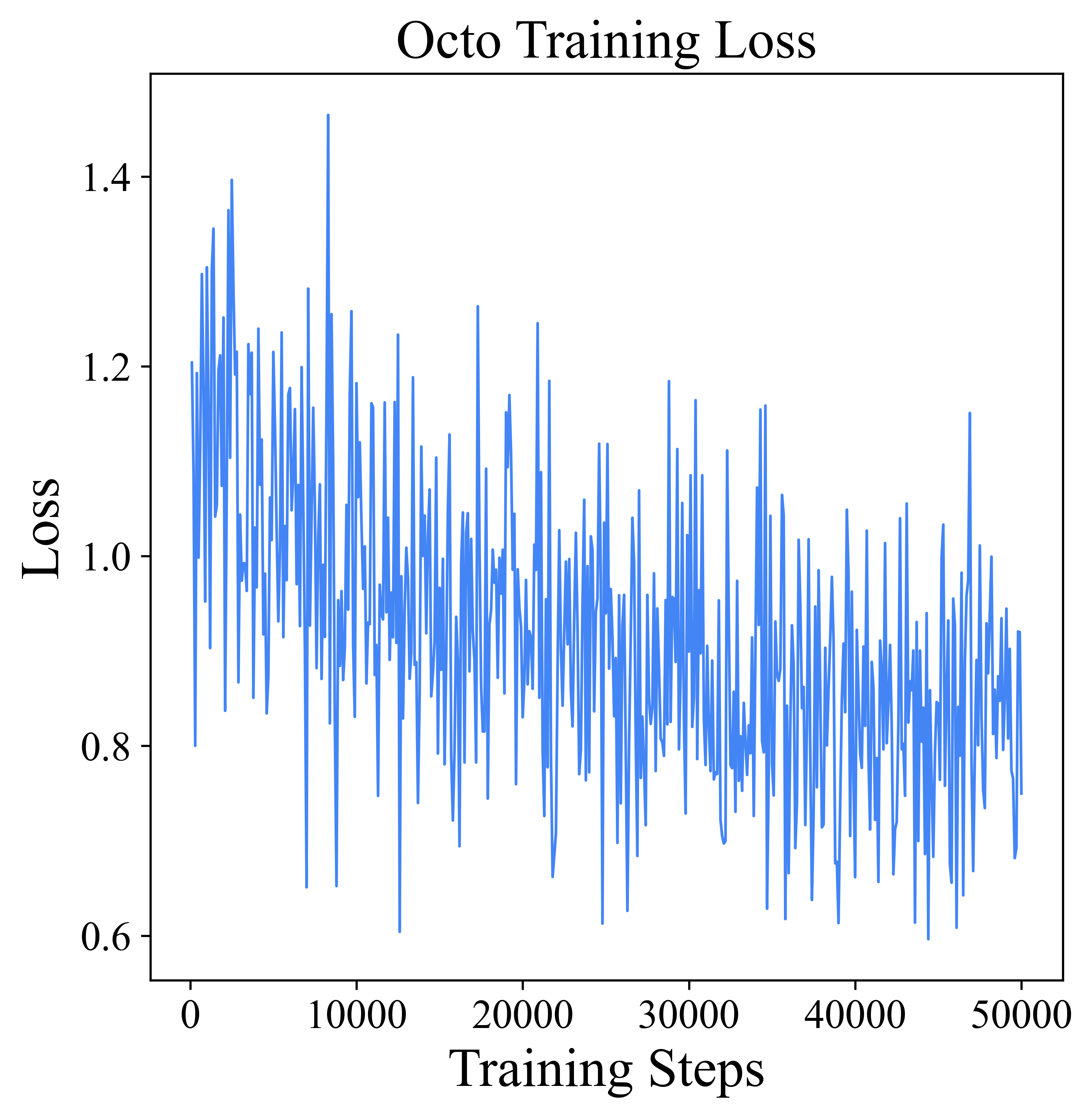}
 \caption{\small The learning curves for OpenVLA and Octo in LIBERO fintuning.}
 \label{fig:openvla_octo_libero}
\end{figure*}

\noindent{\textbf{Baseline models}}.
We include 6500 expert demonstrations of 130 different tasks collected with Franka Robot in LIBERO~\cite{liu2024libero} simulation to train our UniAct-0.5B and then follow the LIBERO Benchmark~\cite{liu2024libero} to evaluate our models. The input images are rendered by the emulator and we use the default $128 \times 128$ resolution. As all the open-source baseline models were not initially trained with the simulation data, substantial effort was put into fine-tuning them to facilitate fair comparisons with our UniAct framework. The fine-tuning process for OpenVLA, Octo and CrossFormer was conducted on 8 A6000 GPUs and 2 4090 GPUs, lasting 7 hours and 4 hours, respectively. Details of the training hyperparameters are provided in Table~\ref{tab:finetune-baseline}. Notably, we manually cleaned the ``no-op'" data before fine-tuning OpenVLA following its official guidance, a step that proved crucial for achieving convergence. However, as shown in Figure~\ref{fig:openvla_octo_libero}, even with an increased number of training steps, the model's action accuracy remained low. In contrast, our UniAct framework demonstrated robustness to noisy data. For LAPA, we employed an identical training recipe to that used for the WidowX robots. The OXE-pretrained LAPA model was fine-tuned with 2,000 gradient steps on Libero, utilizing a batch size of 64.

\noindent{\textbf{Adapt to ACT head}}. To evaluate UniAct's adaptability to heterogeneous decoder heads with varying model architectures, we augmented UniAct-0.5B with an additional ACT head and fine-tuned it on the Libero dataset. Leveraging ACT's sophisticated architectural design, we incorporated image observations from a wrist-mounted perspective and proprioceptive data as supplementary inputs to the decoder head. We adhered to the official implementation settings for the ACT model. Fine-tuning was conducted for 200K gradient steps, utilizing a batch size of 256 across 8 A100 GPUs.

\begin{table}[h]
    \centering
    \resizebox{0.45\textwidth}{!}{
    \begin{tabular}{l|c|c}
        \toprule
        Model-Task & Config & Value \\
        \midrule
        \multirow{6}{*}{\centering OpenVLA-LIBERO} 
        & Optimizer & AdamW  \\
        & Batch size & 64 \\
        & Learning rate & $3\times10^{-4}$ \\
        & Weight decay & $1\times10^{-2}$ \\
        & Optimizer momentum & $\beta_1=0.9, \beta_2=0.999$ \\ \midrule
        \multirow{6}{*}{\centering Octo-LIBERO} 
        & Optimizer & AdamW  \\
        & Batch size & 64 \\
        & Learning rate & $3\times10^{-4}$ \\
        & Weight decay & $1\times10^{-2}$ \\
        & Optimizer momentum & $\beta_1=0.9, \beta_2=0.999$ \\
        &Warmup iters & 2000 \\ \midrule
        \multirow{6}{*}{\centering OpenVLA-AIRBOT} 
        & Optimizer & AdamW  \\
        & Batch size & 64 \\
        & Learning rate & $3\times10^{-4}$ \\
        & Weight decay & $1\times10^{-2}$ \\
        & Optimizer momentum & $\beta_1=0.9, \beta_2=0.999$ \\ \midrule
        \multirow{6}{*}{\centering Octo-AIRBOT} 
        & Optimizer & AdamW  \\
        & Batch size & 64 \\
        & Learning rate & $3\times10^{-4}$ \\
        & Weight decay & $1\times10^{-2}$ \\
        & Optimizer momentum & $\beta_1=0.9, \beta_2=0.999$ \\
        &Warmup iters & 2000 \\ 
        \bottomrule
    \end{tabular}
}

    \caption{Fine-tuning hyper-parameters for baseline models in simulation.}
    \label{tab:finetune-baseline}
\end{table}

\subsection{Fast adaptation to AIRBOT}

To assess the fast adaptation capability of UniAct-0.5B and baselines, we fine-tune them  using newly collected demonstrations on an unseen embodiment during the pretraining phase, AIRBOT. In this section, we provide detailed information about the fine-tuning processes.

\noindent{\textbf{Fine-tuning Settings For UniAct}}:
We utilized 4 A100 GPUs to fine-tune UniAct-0.5B with DeepSpeed. Notably, we train a new MLP network from scratch as the heterogeneous head for AIRBOT while keeping the other modules frozen. The fine-tuning was conducted over a span of 1 hours. The training hyper-parameters employed during this process are detailed in Table~\ref{tab:airbot-uniact}.

\begin{table}[h]
    \centering
    \begin{tabular}{l|c}
        \toprule
        config & value \\
        \midrule
        optimizer & AdamW  \\
        batch size &  128 \\
        learning rate & $3\times10^{-4}$ \\
        weight decay & 0. \\
        optimizer momentum & $\beta_1,\beta_2$=0.9,0.95  \\
        iters & 10K \\
        model precision & BFloat16 \\
        \bottomrule
    \end{tabular}
    \caption{Hyper-parameters for fine-tuning UniAct-0.5B on AIRBOT}
    \label{tab:airbot-uniact}
\end{table}

\begin{figure*}[t]
    \centering
    \includegraphics[width=0.3\textwidth,height=0.25\textwidth]{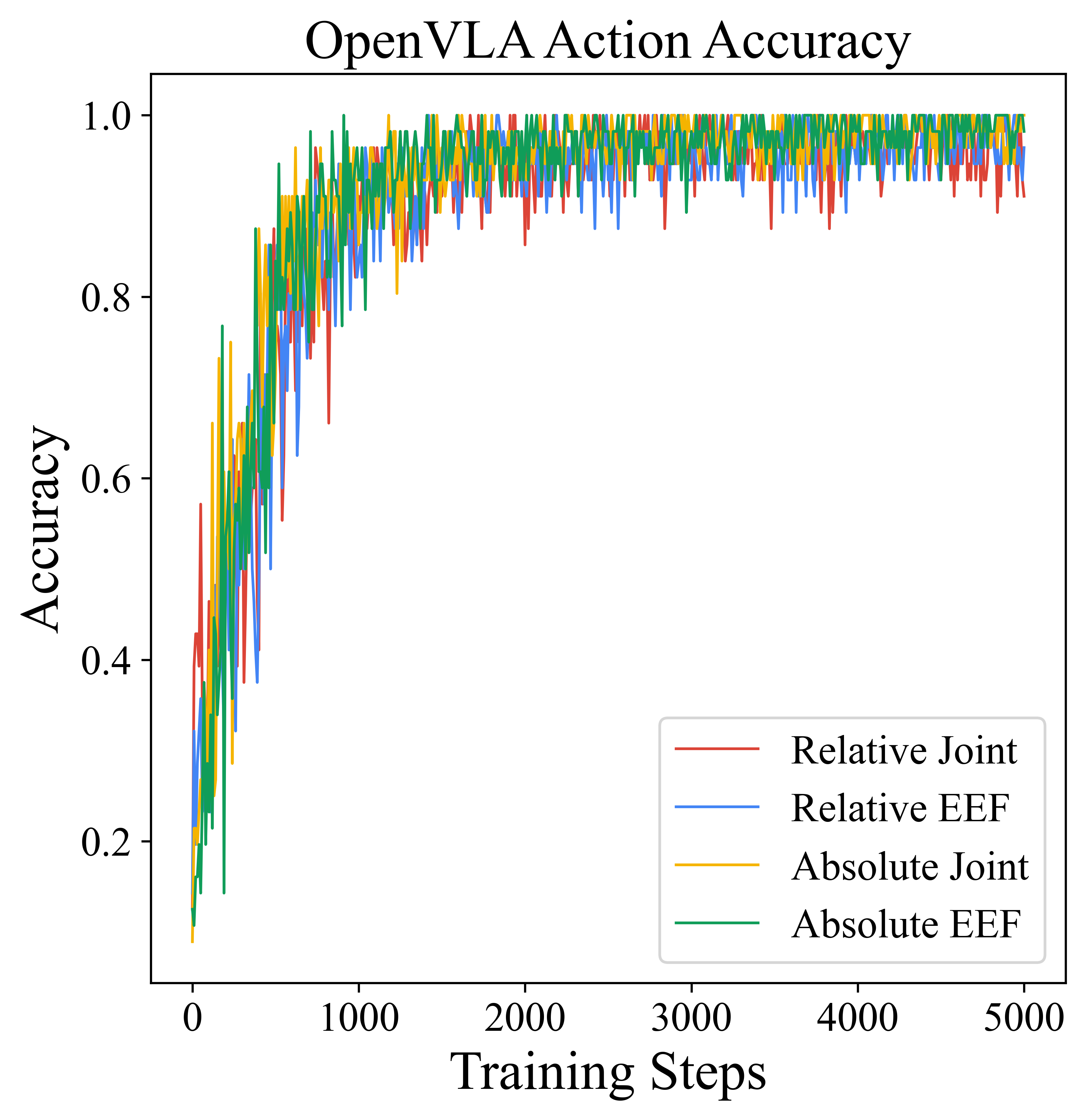}
    \includegraphics[width=0.3\textwidth,height=0.25\textwidth]{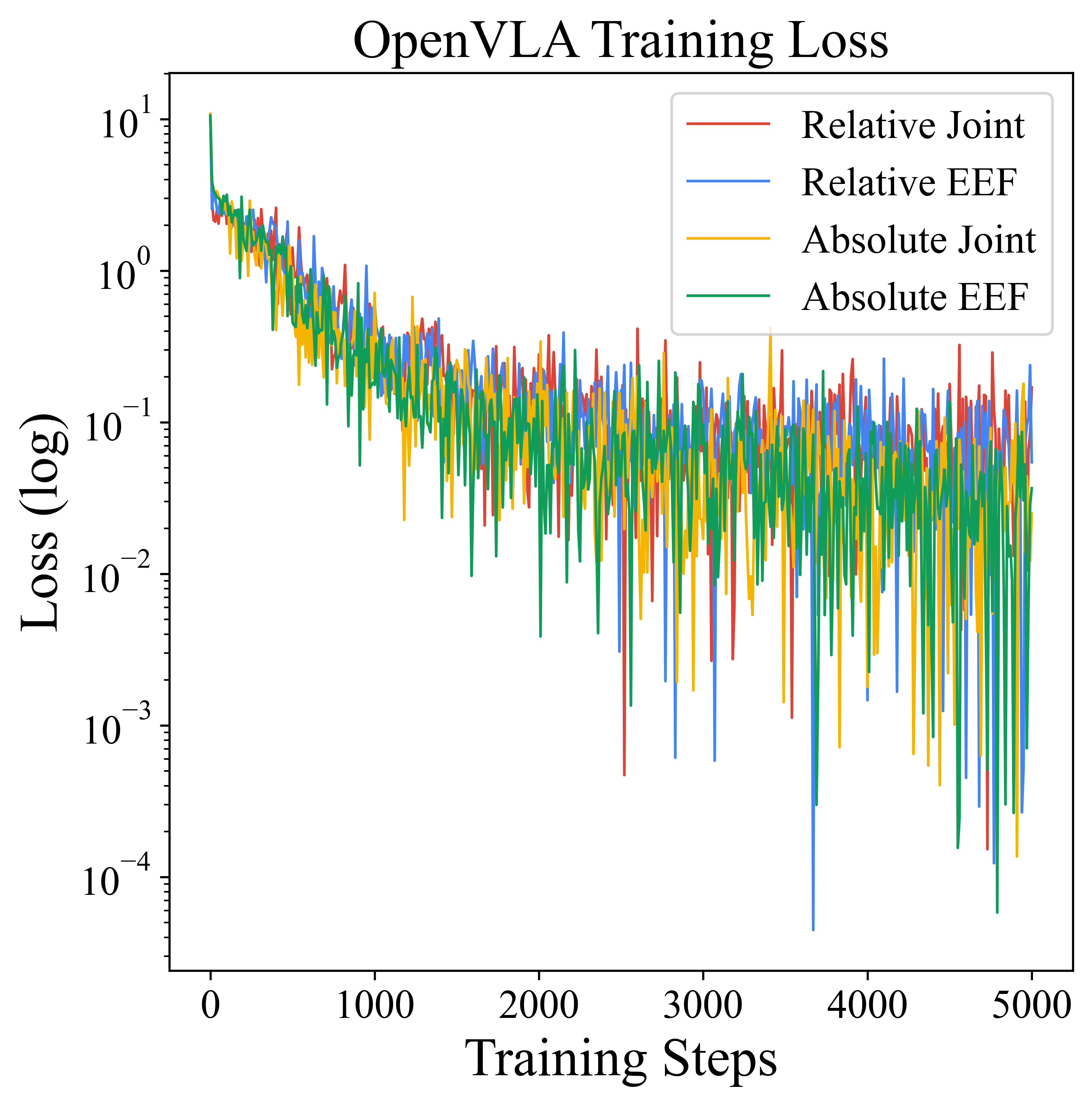}
    \includegraphics[width=0.3\textwidth,height=0.25\textwidth]{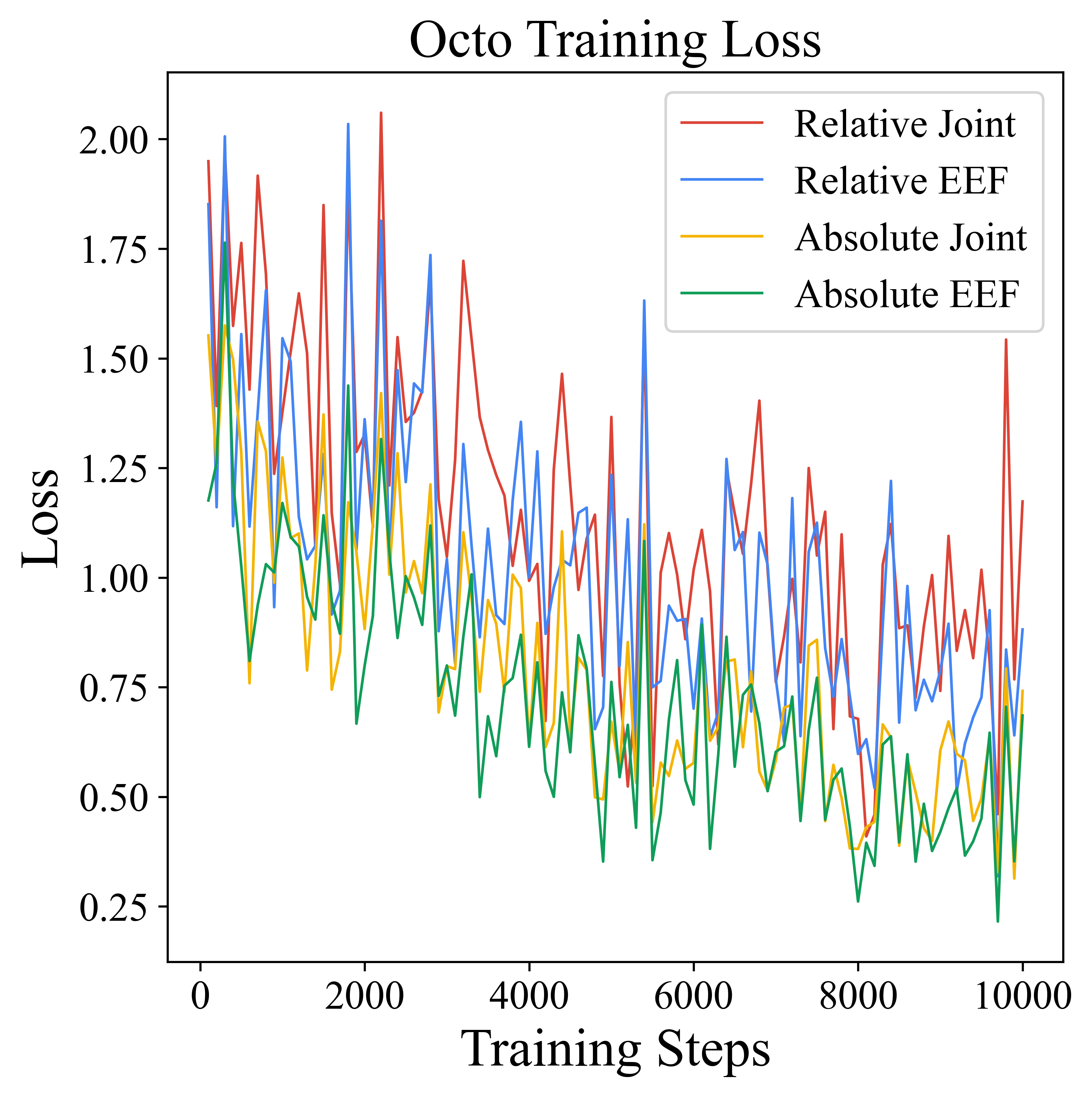}
 \caption{\small The learning curves for OpenVLA and Octo in AIRBOT finetuning.}
 \label{fig:openvla_octo_airbot}
\end{figure*}

\noindent{\textbf{Fine-tuning Settings For Baseline Models}}. The fine-tuning for the baseline models, OpenVLA and Octo, was executed on 8 A6000 GPUs and 2 4090 GPUs, lasting 1.5 hours and 0.5 hours, respectively. To ensure robust performance, the training of OpenVLA continued until it reached an accuracy rate of 95\% on the training dataset, aligning with the official requirements in~\cite{kim2024openvla}. Training hyper-parameters for this process are illustrated in Tab~\ref{fig:openvla_octo_airbot}.

\subsection{Fast adaptation to Bi-manual AIRBOT}
~\label{sec:fast-adapt-to-bimanual}
To assess UniAct's versatility across diverse embodiments, we performed fast adaptation experiments using the bi-manual AIRBOT. Given the requirement for precise perception of complex environments inherent in bi-manual robotic manipulation, which a simple MLP decoder head cannot fulfill, we employed an ACT decoder head for the bi-manual AIRBOT. We designed four challenging manipulation tasks for evaluation: 'sweep plate', 'fold towel', 'put cup on plate', and 'transport pen'. For each task, we collected 250 demonstration trajectories and subsequently fine-tuned UniAct-0.5B. Training hyperparameters are detailed in Tab~\ref{tab:2airbot-uniact}

\begin{table}[h]
    \centering
    \begin{tabular}{l|c}
        \toprule
        config & value \\
        \midrule
        optimizer & AdamW  \\
        batch size &  256 \\
        learning rate & $5\times10^{-4}$ \\
        weight decay & 0. \\
        optimizer momentum & $\beta_1,\beta_2$=0.9,0.95  \\
        iters & 100K \\
        model precision & BFloat16 \\
        \bottomrule
    \end{tabular}
    \caption{Hyper-parameters for fine-tuning UniAct-0.5B on Bi-manual AIRBOT}
    \label{tab:2airbot-uniact}
\end{table}

\section{More Related Works}
\label{sec:related_appendix}

\noindent\textbf{Embodied Models with Hierarchical Structures}. UniAct resembles a hierarchical-style structure, which firstly infers an universal action and then translate it into actionable actions. Some works also explore hierarchical structures, utilizing the planning capabilities of LLMs or VLMs to decompose original instructions into linguistic~\cite{shi2024yell, ahn2022can, Song_2023_ICCV, zawalski2024robotic, mu2023embodiedgpt} or visual~\cite{blackzero, duvideo, wen2023any} sequences of subgoals. However, transferring these language or visual subgoals into precise actions still requires extensive action labels and struggles to leverage the shared structures across diverse action spaces. The universal action space in UniAct, however, serves as a more fine-grained skill library that can be efficiently adapted to physically grounded actions. This space is also end-to-end trained, maximizing the VLMs’ capabilities to develop a flexible and comprehensive structure.

\section{Limitations and Future Works}
\label{sec:limit_appendix}
In this section, we discuss limitations in this work and the corresponding solutions. We hope this will inspire more interesting works.

\noindent{\textbf{More Embodiments}}:
In this study, we explore the concept of a universal action space that can be shared across different action spaces. In this current version, UniAct is mostly evaluated with varied control interfaces on single robotic arms. The underlying motivation stems from an intuitive observation: despite differences in control interfaces, these robotic arms inherently exhibit common physical movements. However, this raises a crucial question: Is the commonality of physical movement exclusive to robotic arms or similar embodiments?

We argue that the answer is no. Future work will explore the universal actions for more complex embodiments, such as dual robotic arms, dexterous robotic hands, quadrupeds and even autonomous driving cars, which differ in degrees of freedom and mechanical structures. Despite their strong heterogeneity, these systems may also share some fundamental movements with simpler robotic arms, holding the potential to be incorporated into the same universal action space. Therefore, we hope to develop a ``truly'' universal action space that is capable of encoding movements across ANY physical embodiment while recognizing unique characteristics and prioritizing shared commonalities.
This offers a promising direction for future research, with strong potential for enabling cross-embodiment control across diverse systems.

\noindent{\textbf{More Flexible Network Design}}:
In our current implementation, UniAct utilizes identical decoding heads—a simple MLP network—for each embodiment to minimize the risk of over-fitting and facilitating training for the universal action extractor. However, it seems more reasonable that the complexity and number of parameters in the decoding heads should be tailored according to the control complexity of each embodiment and the diversity of its training data. For example, a bi-manual robot, with its increased degrees of freedom, would logically require a more complex decoding head than a single-arm robot to effectively learn and decode universal actions. 

Looking forward, as the dataset expands to include more demonstrations from more diverse embodiments, it will become crucial to develop specialized decoding heads that consider the specific control complexities of each embodiment. Additionally, incorporating embodiment-specific information, such as proprioceptive data or different views, into these decoding heads could significantly enhance their performance.


\noindent{\textbf{Scaling Law For Universal Action Training}}:
While our efforts to train UniAct with a vast array of open-source data have yielded commendable results, an intriguing question has arisen: Does more data unequivocally improve the universal action space? This question calls for a thorough examination from multiple perspectives. Firstly, does the incorporation of more embodiments inherently enhance the action space, or is there greater value in accumulating diverse task demonstrations for the same embodiment? Moreover, it's critical to consider whether the more data always bring better result.

Studying the relationship between the amount of data and how well our universal action space performs could be very useful. By understanding these relationships, we can better plan our data collection to improve model training effectively and efficiently.

\noindent{\textbf{More Utilization of Universal Action}}. Our work has highlighted the robust capabilities of universal actions in deploying cross-embodiment robot policies. However, we believe that the potential applications of universal actions extend far beyond what has been explored so far.

One promising direction is in the development of world models, where universal actions serve as a form of 'tokenizer'. This role involves breaking down complex actions into standardized, understandable components. By employing a universal action space for planning, these world models can more accurately predict and simulate outcomes across various scenarios and environments. This uniform approach to understanding and interacting with the world is invaluable for advanced planning and decision-making in robotics.


\end{document}

%% file: preamble.tex
\usepackage{tabularx, multirow}

%% file: sec/0_abstract.tex
\begin{abstract}
\vspace{-8pt}


Training on diverse, internet-scale data is a key factor in the success of recent large foundation models. Yet, using the same recipe for building embodied agents has faced noticeable difficulties. Despite the availability of many crowd-sourced embodied datasets, their action spaces often exhibit significant heterogeneity due to distinct physical embodiment and control interfaces for different robots, causing substantial challenges in developing embodied foundation models using cross-domain data. In this paper, we introduce UniAct, a new embodied foundation modeling framework operating in a \textbf{\underline{Uni}versal \underline{Act}ion Space}. Our learned universal actions capture the generic atomic behaviors across diverse robots by exploiting their shared structural features, and enable enhanced cross-domain data utilization and cross-embodiment generalizations by eliminating the notorious heterogeneity. The universal actions can be efficiently translated back to heterogeneous actionable commands by simply adding embodiment-specific details, from which fast adaptation to new robots becomes simple and straightforward. Our 0.5B instantiation of UniAct reaches 14X larger SOTA embodied foundation models in extensive evaluations on various real-world and simulation robots, showcasing exceptional cross-embodiment control and adaptation capability, highlighting the crucial benefit of adopting universal actions. 
Project page: 
\href{https://2toinf.github.io/UniAct/}{\texttt{https://2toinf.github.io/UniAct/}}

\end{abstract}

%% file: sec/1_intro.tex
\vspace{-10pt}
\section{Introduction}
\label{sec:intro}
\vspace{-3pt}

In fields like natural language processing and computer vision, foundation models trained on vast and diverse data sources have demonstrated remarkable success and strong generalization ability, highlighting the benefits of learning general-purpose models over task-specific counterparts
~\cite{achiam2023gpt, liu2024visual, bai2023qwen, touvron2023llama}. Inspired by these successes, developing versatile embodied foundation models that are capable of handling cross-task, cross-environment, and cross-embodiment generalization,
offers a promising pathway towards building general-purpose embodied agent~\cite{kim2024openvla, yang2024pushing, doshiscaling, team2024octo, driess2023palm, jang2022bc, black2024pi0visionlanguageactionflowmodel}.

\begin{figure}[t]
    \centering
    \includegraphics[width=0.9\linewidth]{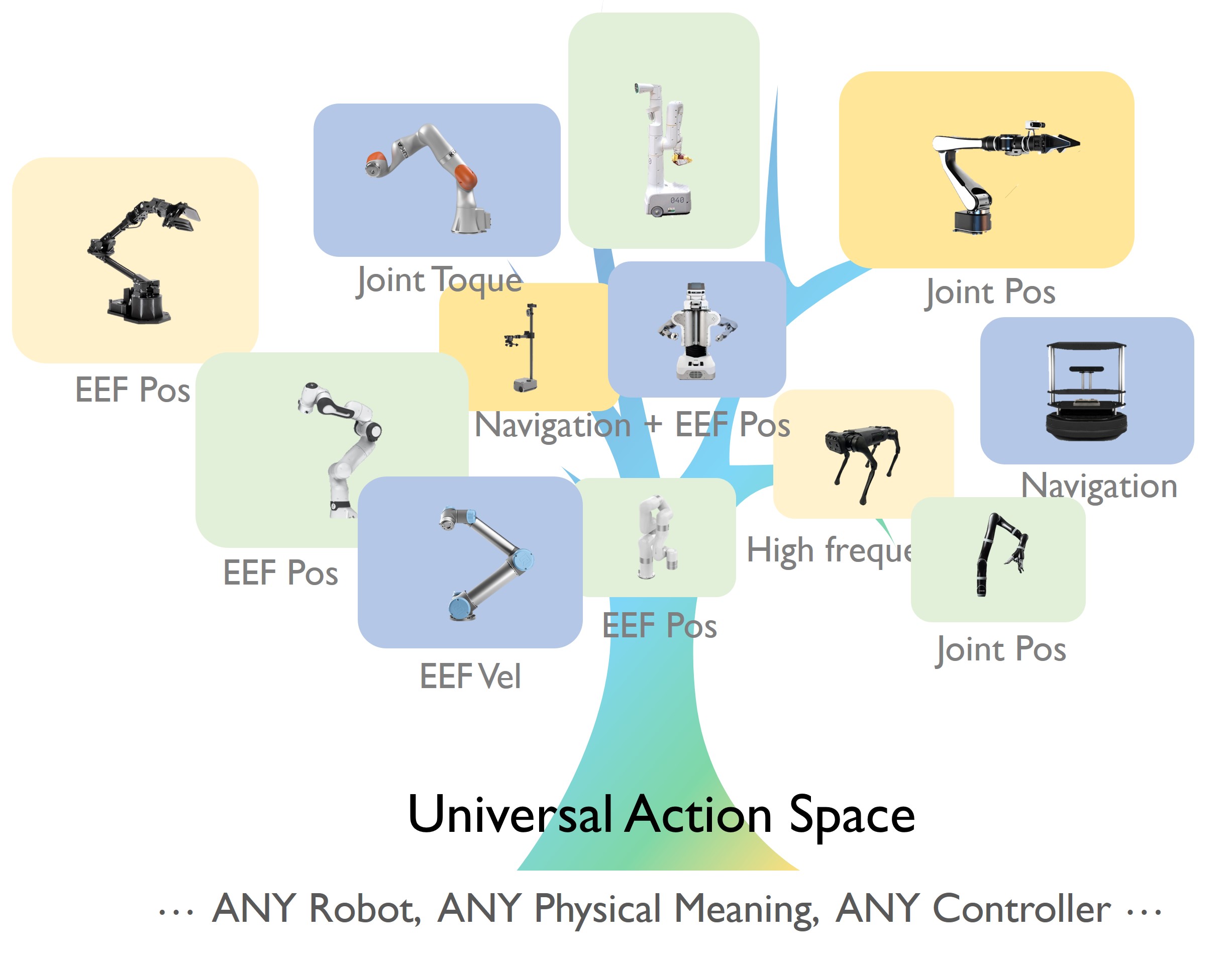}
    \vspace{-10pt}
    \caption{Universal action space holds the potential to be versatile to ANY domain-specific actions.}
    \vspace{-15pt}
    \label{fig:enter-label}
\end{figure}

However, significant challenges arise from the substantial heterogeneity of embodied data 
~\cite{o2023open, fang2024rh20t, walke2023bridgedata}.
Such heterogeneity is evident not only in visual discrepancies caused by variations in camera placements (\textit{e.g.}, wrist or third-person view) and environmental conditions (\textit{e.g.}, lighting or background variations), but more critically, in 
\textit{action heterogeneity}~\cite{doshiscaling, hpt_arxiv, niu2024comprehensive}. \textit{1)} Robots with different embodiment (\textit{e.g.}, different degrees of freedom or distinctions across robotic arms, quadrupeds, and cars)  possess entirely distinct action spaces~\cite{yang2024pushing}. \textit{2)} Furthermore, the diversity in control interfaces (\textit{e.g.}, end effector (EEF) position or velocity controller for robotics arms) leads to fundamentally different physical meanings for action commands~\cite{o2023open}. \textit{3)} Even when actions are collected from the same robotic platform but by different human manipulators, the multi-modality in human behaviors also exacerbates such heterogeneity~\cite{liu2024rdt, chi2023diffusion, bet, vqbet}. Consequently, embodied action data collected across different robots and institutions
tend to reside on largely disjoint manifolds within the original physical spaces (e.g., position and rotation of end-effectors)~\cite{doshiscaling, hpt_arxiv}, significantly complicating data sharing across different data sources.

Currently, no existing solution can adequately address the issue of action heterogeneity. Most prior studies forcibly treat different action spaces as equivalent and apply the same discretization or normalization techniques, leading to a potentially 
conflicted action space where similar action encodings could represent entirely different physical meanings \cite{team2024octo, kim2024openvla, o2023open}. While some efforts attempt to 
design a physically interpretable action space that is applicable across various robotic systems by na\"ively aggregating
all individual action spaces~\cite{liu2024rdt, doshiscaling}. This requires extensive human engineering efforts and fails to uncover and leverage the inherent connections across different embodied action spaces,
impeding the effective development of general-purpose embodied foundation models. 


In this paper,
we introduce \textit{UniAct} (Embodied foundation models with \underline{Uni}versal \underline{Act}ions), a novel embodied modeling framework that is constructed in the \textit{Universal Action Space} rather than the troublesome heterogeneous action spaces. 
The universal actions learned in \textit{UniAct} encode the generic atomic behaviors that are generalizable across diverse robotics platforms without being constrained to specific
embodiment mechanics and control interfaces. 
For instance, different robots should perform similar behaviors of ``moving forward" when facing a target directly ahead, despite exhibiting totally different control signals due to their embodiment gaps. 
This abstraction transcends specific embodiment and control constraints, allowing it to be universally applicable across diverse robotics platforms, providing significant potential to enhance cross-embodiment data utilization and model generalization. In this sense, minimal parameters and data are sufficient to decode the universal action to an embodiment-specific one, since the general motion behaviors have been captured in the universal space and the decoder simply needs to add some embodiment details for each robotic platform, therefore enabling efficient adaptation across new robotic systems during deployment.

In details, UniAct employs a shared Vision Language Model (VLM)~\cite{li2024llava, liu2024visual, liu2024improved, bai2023qwen_vl, zheng2024instruction, lai2024lisa} to construct the universal action space as a vector-quantized codebook~\cite{van2017neural}.
Akin to a learnable skill library~\cite{bet, vqbet, mete2024quest}, each code encapsulates an atomic behavior versatile enough to be executed by diverse robots. This setup acts as a crucial information bottleneck, driving the VLM to identify and exploit shared primitive behaviors across diverse action spaces.
This extraction scheme enables effective generalization of behaviors for cross-embodiment control, making our 0.5B instantiation, UniAct-0.5B, surpasses models 14x larger, such as OpenVLA~\cite{kim2024openvla} with 7B parameters, in a wide range of tasks.
The derived universal actions can be translated into precise, actionable commands for various embodiments through streamlined heterogeneous decoders. These decoders take the universal action as conditional input and augment it with embodiment-specific features from their unique observational data. This allows for flexible customization according to specific requirements, such as the inclusion or exclusion of proprioceptive features or variations in the number of camera views. Fast adaptation to new domains or robotic platforms can be achieved by simply adding new heads as lightweight decoders for new tasks.
Our comprehensive evaluations on challenging task settings, including large view changes and an unseen robot not present in the training data, confirm UniAct's remarkable transferability, demonstrating great advantages of developing embodied foundation models within a universal action space over the conventional heterogeneous spaces.

\begin{figure*}[t]
    \centering
    \includegraphics[width=0.95\linewidth, height=9.5cm]{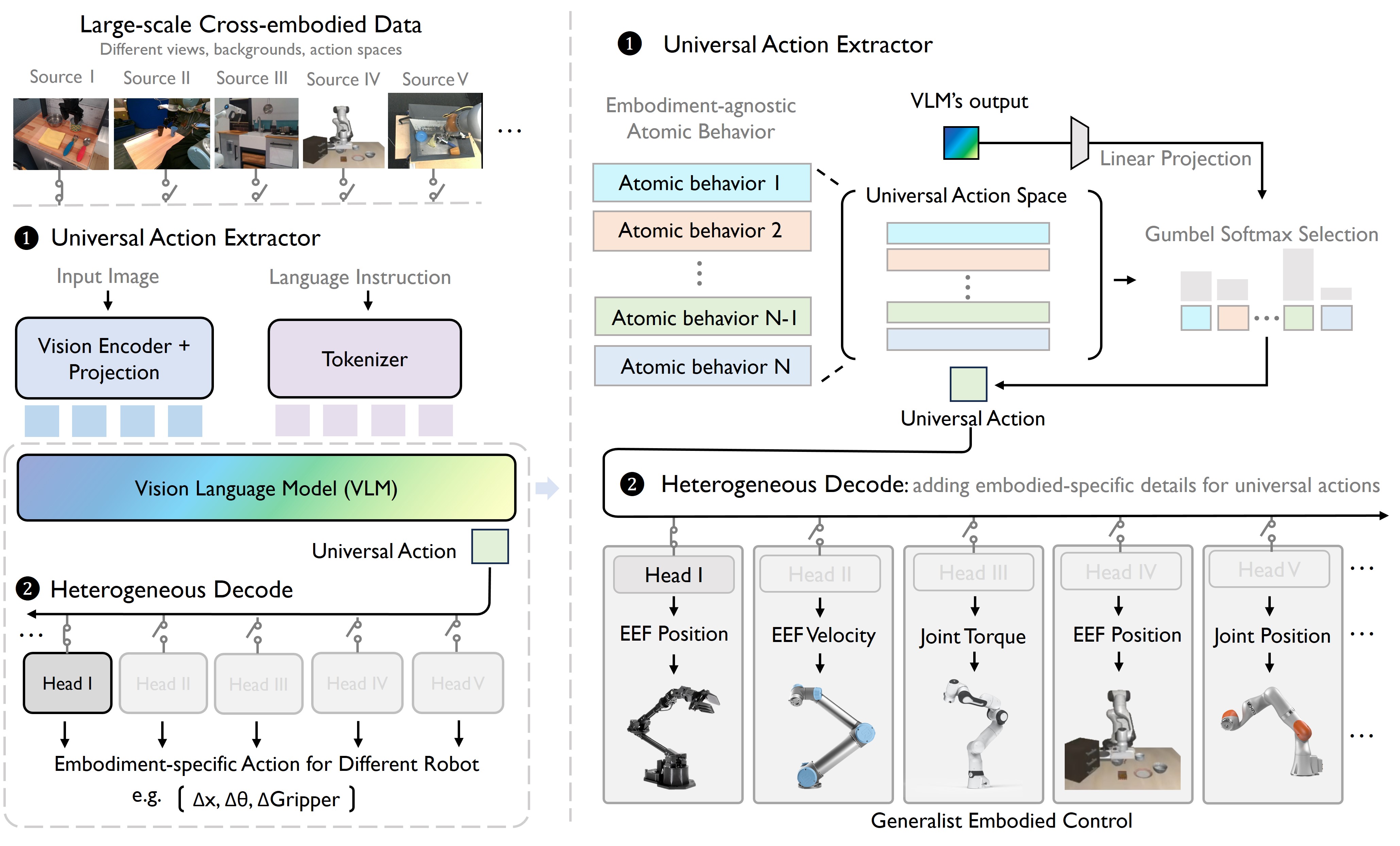}
    \vspace{-8pt}
    \caption{\small 
    UniAct leverages diverse data sources for generalist embodied control. A shared Vision Language Model (VLM) extracts transferable features across various data sources; The output tokens are converted into universal actions, represented as vector-quantized codes where each code captures common atomic behaviors across different robots; The Gumbel Softmax-selected universal action is then translated back into specific commands through different heads, each encoding embodiment-specific features for individual robots. 
    }
    \label{fig:intro}
    \vspace{-10pt}
\end{figure*}

\vspace{-5pt}
\section{Related Work}
\vspace{-5pt}

\textbf{Multimodal Foundation Models.} Large Language Models (LLMs)~\cite{achiam2023gpt, liu2024visual, bai2023qwen, touvron2023llama} have exhibited remarkable capabilities across a variety of tasks, showcasing impressive zero-shot and in-context learning capabilities~\cite{dong2022survey}. Building on this, large Vision Language Models (VLMs) have been developed by integrating vision and language into a unified tokenized space, demonstrating outstanding multimodal instruction-following abilities~\cite{li2024llava, bai2023qwen_vl, lai2024lisa, zheng2024instruction, sun2023emu, sun2024generative, fei2022towards}. Their success is primarily attributed to extensive internet-scale pretraining, which leverages vast and diverse high-quality data corpora from the Internet.

\noindent\textbf{Embodied Foundation Models.}
When developing embodied foundation models, an additional crucial modality—action (the deployable control signals that robots can interpret and execute, \textit{e.g.}, EEF position/velocity)—is incorporated during training. State-of-the-art models are often constructed as Vision Language Action models (VLA)~\cite{driess2023palm, kim2024openvla, brohan2022rt, brohan2023rt, o2023open}, integrating visual and linguistic inputs with actionable outputs.
However, the action labels collected from different robotics platforms and labs exhibit significant heterogeneity~\cite{o2023open, walke2023bridgedata, khazatsky2024droid}, impeding
effective data sharing across different sources. To sidestep this challenge, many works employ
large-scale action-free vision language data, such as out-of-domain human activities~\cite{damen2018scaling, grauman2022ego4d, goyal2017something}, to firstly obtain a good embodied VLMs, then finetune it to a specialized VLA given a small set of action labels from a specific robot platform~\cite{wuunleashing, cheang2024gr, karamcheti2023language, ma2023liv, ye2024latent, anonymous2024igor, lidecisionnce, li2024robo, zheng2024instruction, nairr3m, brohan2023rt, ajay2024compositional, duvideo, blackzero, wen2023any, gao2024physically}. While these methods can enhance sample efficiency for specific robots on a narrow set of tasks, they suffer from serious performance bottlenecks towards building a generalist embodied agent
~\cite{shi2024yell, duvideo, blackzero}, 
as the action data gathered from any single robot platform is far less comprehensive than crowed-sourced data collected globally~\cite{o2023open, khazatsky2024droid, fang2024rh20t}.

Some recent works leverage the abundant heterogeneous action labels to develop generalist robot policies for cross-embodiment control. RT-X series~\cite{o2023open}, Octo~\cite{team2024octo} and OpenVLA~\cite{kim2024openvla} leverage data from different 7-DoF robots to enhance generalization over the one trained on single robot source. Step further, CrossFormer~\cite{doshiscaling}, RDT~\cite{liu2024rdt}, $\pi_0$~\cite{black2024pi0visionlanguageactionflowmodel} and ~\citet{yang2024pushing} explore the data from robots with totally distinct mechanical structures, such as those in manipulation and navigation, and from single-arm versus bi-manual systems. However, existing works either ignore the heterogeneous properties of action spaces of different sources, crudely treating them as equal without considering their inherent conflicts~\cite{kim2024openvla, team2024octo, o2023open}, or na\"ively aggregate all action spaces together, failing to exploit the underlying shared commonalities across different robots~\cite{liu2024rdt, black2024pi0visionlanguageactionflowmodel, doshiscaling, yang2024pushing}.

\noindent\textbf{Embodied Models with Latent Action Spaces.} Our work aims to extract a versatile universal action space, akin to a latent space but encodes common atomic control behaviors and patterns across various robotic platforms. Some works develop embodied models in latent spaces~\cite{chen2022lapo, vqbet, bet, mete2024quest, yuan2024pre, schmidtlearning, ye2024latent, anonymous2024igor}. Among them, LAPA~\cite{ye2024latent}, IGOR\cite{anonymous2024igor}, and LAPO~\cite{schmidtlearning} develop a latent action space through joint self-supervised training of inverse and forward dynamics models on action-free videos~\cite{bruce2024genie}. However, the latent actions extracted in this way primarily focus on explaining the changes between video frames, lacking embodiment considerations or direct causal connections to actual control signals. To see why this is problematic, assuming we add a new object in front of the robot, the visual inputs will change but this has nothing to do with the control behavior, an ideal encoded action should not capture such distracted information.
BeT~\cite{bet}, VQ-BeT~\cite{vqbet} and QueST~\cite{mete2024quest} also build a discrete codebook of actions via K-means clustering~\cite{MacQueen1967SomeMF} or Vector Quantization~\cite{van2017neural, lee2022autoregressive, mentzerfinite}, where each code in the codebook encodes a different clustering center for action labels. These works primarily focus on simpler domains with a single embodiment type, which enhances the ability to model complex human demonstrations with multiple modes, but struggles to address action heterogeneity across different embodiments.
In contrast, our universal actions integrate goal information from the embodiment-agnostic language modality with supervision on the actual action signal, providing a versatile and abstracted skill library to facilitate cross-embodiment sharing. Moreover, our research delves into a more complex heterogeneous setting and develops a large embodied foundation model, moving beyond the limited scopes considered in previous studies.

\vspace{-3pt}
\section{The UniAct Framework}
\vspace{-3pt}
We introduce UniAct, an embodied foundation modeling framework designed to operate in a \textit{Universal Action Space}, adept at bridging domain gaps and facilitating training on large-scale heterogeneous data. We first discuss the desirable properties of universal actions, and then provide a detailed discussion about the model architecture and learning scheme for extracting and decoding universal actions from heterogeneous cross-embodiment data.

\vspace{-1pt}
\subsection{Universal Action Space}
\vspace{-1pt}
The desired universal action space
is that all movements, driven by heterogeneous control signals from various embodiments, can be distilled into shared latent atomic behaviors, despite their distinct physical meanings. We refer to these abstract behavior representations as \textit{universal actions},
which are shared across all physical embodiments.

We are particularly interested in exploring a discrete universal action space.
This is motivated by the robust capabilities of discrete representations in complex reasoning, planning, and predictive learning, as demonstrated by the success of LLMs and VLMs~\cite{achiam2023gpt, bai2023qwen, team2024gemma, bai2023qwen_vl} and Vector Quantized Variational Autoencoders~\cite{van2017neural, peebles2023scalable}. 
In this paper, we model the universal action space as $\mathcal{U} \in \mathbb{R}^{N \times D}$ and implement it with a vector quantized codebook~\cite{van2017neural}, represented as
\begin{equation}
    \mathcal{U} = (u_1,u_2,u_3...u_N),
\end{equation}
where $N$ is the space size and each $u_i$ is a $D$-dimensional vector embedding that represents a generic atomic behavior.

Several prior studies~\cite{bruce2024genie,ye2024latent, anonymous2024igor} pursued a similar concept of constructing generic, latent actions
by inferring them as the dynamic changes observed between two visual states. However, this scheme suffers two key limitations, leading to suboptimal and noisy universal action spaces:

\begin{itemize}
\item The observational changes encompass not only the outcomes of robots, but also external factors (e.g., the appearance of new objects, human intervention, etc.) that have no causal connection to the actual control.
\item The interval between two observations critically influences the semantic interpretation of the extracted atomic behaviors, making standardizing behavioral interpretation complicated across different data sources.
\end{itemize}

\vspace{-1pt}
\subsection{Universal Action Extraction}
\vspace{-1pt}
To derive the desirable universal action space, we propose a new 
method for extracting universal actions, pivoting away from solely focusing on explaining observational changes, but more on understanding task progression. 
Specifically, we fine-tune a large vision language model as the universal action extractor, which outputs the likelihood $p(u|o, g)$ of selecting universal action $u$ given observation $o$ and task goal $g$ (e.g., language instruction). We want the corresponding adopted universal action $u^*$ that matches the atomic behavior encoded in the embodied data to satisfy:
\begin{equation}
    u^* = \mathop{\arg \max}_{u\in \mathcal{U}} p(u|o, g)
\end{equation}
Akin to planning in the latent space~\cite{chen2022lapo, park2024hiql, parkfoundation}, the extractor aims to infer the most relevant universal action for solving a given task $g$ under the observation $o$, 
thereby crafting universal actions directly related to task progression rather than merely identifying the noisy observational changes. We employ a VLM to achieve this purpose due to its strong visual-language reasoning capability.
Moreover, fine-tuning a pre-trained VLM also greatly improves the sample efficiency when learning the universal actions.
This extractor creates a crucial information bottleneck for cross-domain generalization, as different robots are forced to use the same 
discrete codebook $\mathcal{U}$ to capture the generic and shared atomic behaviors across all domains. 

To implement this, however, the non-differentiable $\mathop{\arg\max}$ impedes the gradient propagation.
So, we use categorical reparametrization during the training, utilizing the Gumbel-Softmax technique to facilitate gradient estimation~\cite{jang2016categorical}. Specifically, the forward procedure is as follows: 
\begin{equation} 
	u^* = \sum_{i=1}^n w_i{u}_i,
\end{equation}
where $w_i$ is the weight for each universal action $u_i$, computed using the Gumbel Softmax function:
\begin{equation}
    w_i = \frac{\mathrm{exp}((\mathrm{log}\ p (u_i |o, g) + \epsilon_i) / \tau}{
    \sum_{k=1}^{n}
    \mathrm{exp}((\mathrm{log}\ p (u_k |o, g ) + \epsilon_k) / \tau}
\end{equation}
Here, $\epsilon_i$ is a Gumbel noise sampled from the Gumbel distribution, and $\tau$ is the temperature to smooth the probability distribution.
To promote parameter space exploration 
in the early training stage and the stability of model convergence, we gradually decay temperature $\tau$ during the training process.
Our proposed universal action extractor is illustrated 
in Figure~\ref{fig:intro}, please refer to the Appendix~\ref{sec:train_appendix} for more details.

\vspace{-1pt}
\subsection{Heterogeneous Decoding}
\vspace{-1pt}
To efficiently translate the highly abstract behaviors in the universal action space into precise, embodiment-specific control signals, it is crucial to integrate more embodiment detail, such as control type, proprioception, and distinct observations. 
To address this, we introduce a series of lightweight decoder heads to adapt for each type of embodiment, denoted as $\mathcal{H} = (h_1...h_k...h_K)$, $K$ is number of training domains. Each head $h_k$ is specifically designed to learn the mapping from universal action $u^*$ and visual observation $o$ to heterogeneous control signals for the embodiment 
in domain $k$.  The operation of each decoder head $h_k$ can be formulated as:
\begin{equation} 
	{\hat{a}}^{(k)} = h_k (u^*, o),
\end{equation}
where ${\hat{a}}^{(k)}$ is the predicted control signals. As overly complex decoding heads with excessive parameters could overfit to
the data distribution of the target domain, all heterogeneous heads are implemented as simple MLP networks that take the $u^*$ and visual features $o$ extracted by a shared vision backbone as inputs. By keeping the decoding heads lightweight, we ensure that the majority of learning is conducted for the universal actions, thereby maximally improving generalization across different embodiments.

\vspace{-1pt}
\subsection{Training Procedure}
\vspace{-1pt}
The primary learning objective of UniAct is to distill a universal action space that is shared across diverse embodiments, with the critical feature that these universal actions can be precisely translated back into domain-specific control signals. To facilitate this, the model is trained using a comprehensive collection of $K$ heterogeneous datasets $\mathcal{D} = (\mathcal{D}_1...\mathcal{D}_k...\mathcal{D}_K)$.  Each $\mathcal{D}_k$ comprises 
a set of robotic control trajectories, represented as $\tau_i^{(k)} = \{o_{i, t}^{(k)}, a_{i, t}^{(k)}, g_i\}_{1\leq t\leq T} $, where $\tau_i^{(k)}$ is the $i$-th trajectory of maximum length $T$, containing observations, actions, and goals. 
Concretely, UniAct ingests $o$ and $g$ as inputs to predict the universal action $u^*$ using the universal action extractor, which is then mapped along with $o$ to ${\hat{a}}^{(k)}$ using the heterogeneous decoding heads. The overall training objective is as follows:
\begin{equation} 
	\mathop{\min}_{\mathcal{U}, \theta}  \sum_{k=1}^K \mathbb{E}_{a_i\in \tau_i, \tau_i \in \mathcal{D}_k} \mathcal{L}_k({\hat{a}}^{(k)}, a_i^{(k)})
\end{equation}
Here, $\mathcal{L}_k$ is the behavior cloning loss, which is customizable based on 
the nature of the action labels in each dataset, for example, employing Cross-Entropy for discrete actions and MSE, Huber loss, or diffusion loss~\cite{ho2020denoising, chi2023diffusion} for continuous actions. We optimize the above objective to learn both the universal action codebook $\mathcal{U}$ as well as the parameters $\theta$ of the universal action extractor and all heterogeneous decoding heads.
Importantly, while $\mathcal{U}$ and the universal action extractor are concurrently updated throughout each training iteration, the heterogeneous heads $h_k$ are updated based on the domain-specific sampled training batches.
This training strategy mirrors the philosophy in many meta-learning methods~\cite{gordon2018metalearning}, which learns both globally shared parameters that allow adaptation to related tasks, as well as task-specific components that guarantee downstream task performance.
Through this approach, UniAct strives to refine a robust, adaptive universal action space as well as a decoding strategy that can be seamlessly integrated with diverse embodiments and their specific operational contexts.

\vspace{-3pt}
\section{Experiments}
\vspace{-3pt}
In this section, we first describe the detailed implementation of the UniAct framework and then present the evaluation experiments conducted to answer the following questions: 

\begin{itemize} 
\item Can universal actions enhance execution performance across various embodiments with large domain gaps? 
\item Can universal actions be seamlessly transferred to new, unseen embodiments? \item Dose UniAct learn a meaningful universal action space? 
\end{itemize}

\begin{figure*}[t]
    \centering
    \includegraphics[width=0.95\linewidth, height=5.2cm]{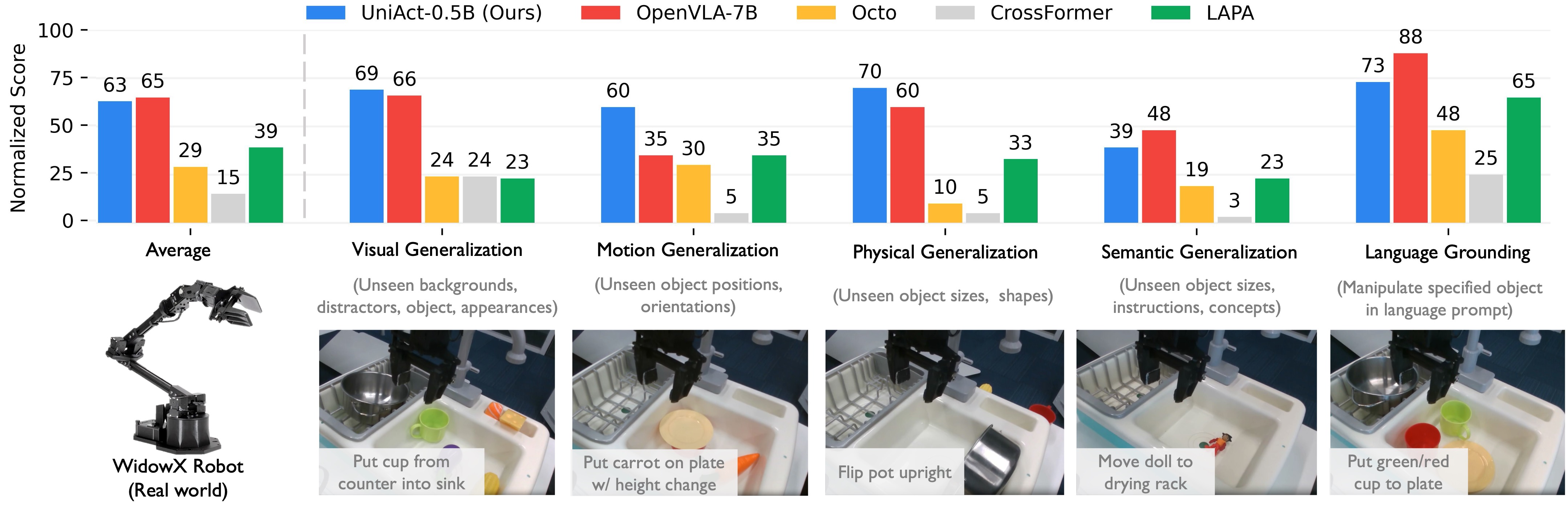}
    \vspace{-5pt}
    \caption{\small WidowX robot evaluation. We evaluate different axes of generalization ability, covering a total of 19 tasks. Representative tasks are presented above. For each task, we evaluate 10 trials and report the average success rate for each suites.}
    \label{fig:widowx_results}
    \includegraphics[width=0.95\linewidth, height=5.2cm]{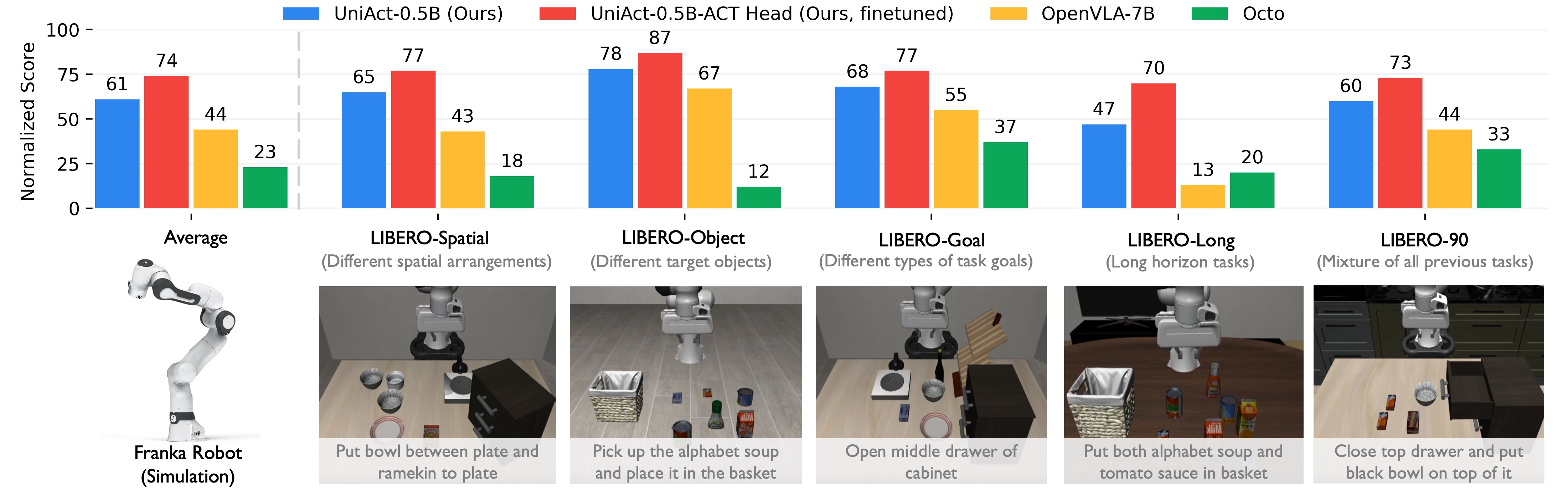}
    \vspace{-7pt}
    \caption{\small For benchmarking purpose, we evaluate on the LIBERO benchmark~\cite{liu2024libero} in simulations. LIBERO contains several task suites to examine different axes of policy abilities. LIBERO-90 contains 90 tasks and others each contain 10 tasks, covering 130 tasks. For each task, we evaluate 20 trials for each model and report the average success rate for each task suites.}
    \label{fig:libero_results}
    \vspace{-0.2cm}
\end{figure*}

\vspace{-1pt}
\subsection{Experiments Setup}
\vspace{-1pt}
\noindent{\textbf{Implementation Details}}. In this paper, we build a 0.5B instantiation of UniAct on heterogeneous embodied data sources to explore a universal action space ($\mathcal{U}\in\mathbb{R}^{256 \times 128}$). Specifically, UniAct-0.5B is built upon \textit{LLaVA-OneVion-0.5B}~\cite{li2024llava}, a well-trained VLM which can provide comprehensive multi-modal representations. The training of UniAct-0.5B is carried out on 64 A100 GPUs with DeepSpeed~\cite{rasley2020deepspeed} over a span of 10 days, utilizing 1 million demonstrations gathered from 28 distinct embodiments. The training data combines several open-sourced robot collections, including Open-X Embodiment~\cite{o2023open}, Libero~\cite{liu2024libero}, and Droid~\cite{khazatsky2024droid}, standardized to include third-person visual observations and language instructions while preserving action heterogeneity. 
For more details about training and data constructions, please refer to the Appendix~\ref{sec:train_appendix}.


\noindent{\textbf{Baseline Setup}}. We select three state-of-the-art (SOTA) vision-language-action models including
 Octo~\cite{team2024octo}, CrossFormer~\cite{doshiscaling}, OpenVLA~\cite{kim2024openvla}, and one SOTA pretrained VLM model, LAPA~\cite{ye2024latent} as baselines. Octo and CrossFormer are 0.1B policies, OpenVLA employs a 7B-parameter auto-regressive architecture with discrete actions. They are trained on about 1 million carefully curated robot demonstrations without action heterogeneity, such as pre-processing all absolute EEF positions to relative EEF positions and removing joint position actions. LAPA is also a 7B model, but instead of using action labels, it constructs latent actions by capturing visual changes between frames based on the same robot demonstrations. To adapt LAPA to generate real actions, further finetuning with action labels is required. 
In contrast, UniAct-0.5B is trained on a similar scale of data from the same data sources but does not employ such tedious data cleaning.
We compare UniAct with the baseline models to 
demonstrate its effectiveness in extracting universal actions from heterogeneous data.

\vspace{-1pt}
\subsection{Main Results}
\vspace{-1pt}
To assess the cross-embodiment generalization capabilities of UniAct-0.5B, we conduct ``out-of-the-box" evaluations on both a real-world WidowX robot \cite{walke2023bridgedata} and a simulation Franka robot from~\citet{liu2024libero}.
Both platforms are commonly used in previous works to test the effectiveness of generalist robot policies~\cite{walke2023bridgedata, o2023open, team2024octo}, and possess substantial domain gaps. Given that our training dataset includes data from these two embodiments, we can leverage the pre-trained heterogeneous heads to translate the universal actions seamlessly
back into deployable control signals.

\noindent{\textbf{Real-World Robot Evaluation}}.
Following~\citet{kim2024openvla}, we define a comprehensive set of evaluation tasks for real-world robots, covering several dimensions of generalization: \textbf{visual}, \textbf{motion}, \textbf{physical}, \textbf{semantic}, and \textbf{language grounding}.
Overall, each model is evaluated across 190 rollouts, distributed over 19 tasks with 10 trials each, see Appendix~\ref{sec:eval_appendix} for more details. Representative tasks and results are illustrated in Figure~\ref{fig:widowx_results}.  To adapt LAPA to WidowX robot, we full-tune it on Bridge~\cite{walke2023bridgedata} dataset about 2K gradient steps. 
UniAct-0.5B outperforms 14X larger OpenVLA-7B and LAPA-7B in visual, motion, and physical generalization tasks. This shows the substantial benefits of extracting universal actions from heterogeneous data in enhancing robustness to visual distractions and low-level control generalization. While OpenVLA leverages a 7B backbone for superior semantic understanding and language grounding capability, UniAct-0.5B achieves comparable performance in semantic generalization and language grounding tasks, underscoring its efficiency and effectiveness.

\begin{figure*}[t]
\centering
\includegraphics[width=0.97\linewidth]{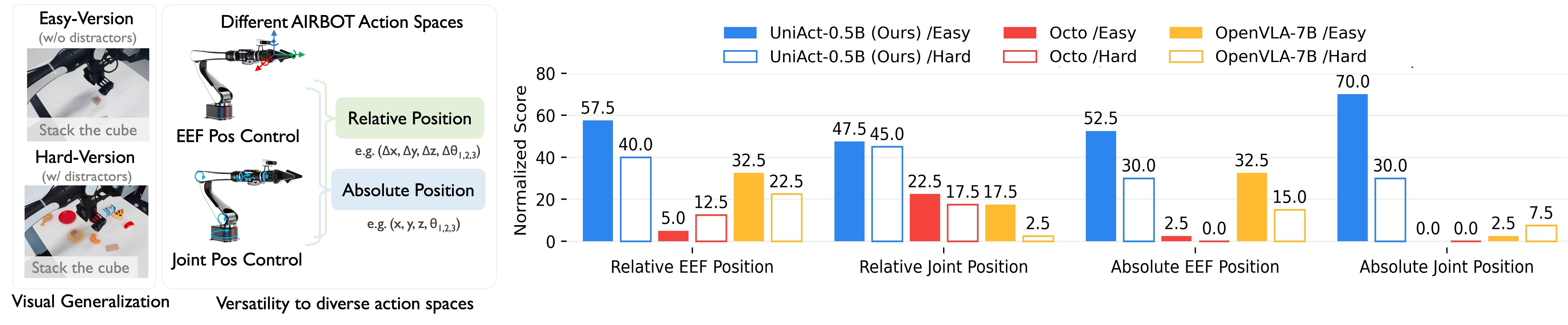}
\vspace{-7pt}
\caption{\small Fast adaptation to new robots. We fine-tune models on an unseen AIRBOT robot with different controller interfaces including relative/absolute joint/EEF positions. The task is to stack the small cube on the small cuboid, requiring precise action execution. The fine-tuning data is collected w/o any distractors, but we also evaluate a hard version, where the policy should be robust to diverse clutters.}
\vspace{-7pt}
\label{fig:airbot}
\end{figure*}

\noindent{\textbf{Simulation Evaluation}}. We utilize the LIBERO Benchmarks~\cite{liu2024libero} for evaluation. Notably, baseline models are not initially trained on the simulation data, thus we fine-tune them on the LIBERO platform. The data used to train both UniAct and the baseline models are fully aligned in terms of tasks, data quantity, and image quality. For more details, please refer to Appendix~\ref{sec:train_appendix}. The benchmark comprises 130 tasks across five suites: LIBERO-Spatial, -Object, -Goal, -Long, and -90. The LIBERO-90 suite includes 90 tasks, while each of the other four suites contains 10 tasks. The results can be found in Fig~\ref{fig:libero_results}. UniAct-0.5B surpasses baseline models in all suites, demonstrating a significant improvement with an overall average accuracy. This superior performance can be attributed to the ability of UniAct to bridge domain gaps and extract generalizable atomic behaviors. By leveraging demonstrations from various domains to learn the universal actions, UniAct significantly enhances task performance on the LIBERO benchmarks.

\vspace{-1pt}
\subsection{Fast Adaptation to New Embodiment}
\vspace{-1pt}

\noindent{\textbf{{Experiment Setup}}}. To assess the fast adaptation ability, we evaluate on a new real robot, AIRBOT (Figure~\ref{fig:airbot}). We evaluate four different controller interfaces: relative/absolute EEF position and relative/absolute joint position. Neither UniAct nor the baselines were pre-trained on AIRBOT data. We collect 100 demonstrations on this new robot platform with the four different types of control interfaces. Considering the significant heterogeneity among these control interfaces, we put a lot of effort into fine-tuning baselines and make sure the model convergence meets the official requirements (e.g., 95\% prediction accuracy for OpenVLA). 

\noindent{\textbf{{Fast Adaptation with UniAct}}}. Unlike baseline models that require extensive training to bridge the adaptation gap across different action types, UniAct can rapidly adapt to new embodiments and control interfaces. Having already learned cross-embodiment behaviors, we facilitate fast adaptation by freezing the codebook and the universal action extractor. Concurrently, we train four heterogeneous decoding heads from scratch for each type of actions with the collected demonstrations. Each newly introduced head is implemented with a simple MLP that takes the $u^*$ and vision features $o$ from a shared vision backbone as inputs.

\noindent{\textbf{Evaluation}}. We conducted evaluations using both an easy and a hard version of the task ``stack the cube on another cube". The results can be found in Figure \ref{fig:airbot}. UniAct-0.5B demonstrates consistently strong generalization across all types of control signals, surpassing the two baseline models. Notably, the number of parameters UniAct-0.5B used for fine-tuning versus the total model size is the smallest (4M / 500M: 0.8\%). In comparison, OpenVLA and Octo utilize 1.4\% (97M / 7000M) and 2\% (2M / 100M) of their total model sizes, respectively. This efficient parameter utilization highlights UniAct's effectiveness and adaptability in applying learned universal actions to new tasks and embodiments with minimal parameter space expansion.

\vspace{-1pt}
\subsection{Fast Adaptation to New Decoder Head}
\vspace{-1pt}

\begin{figure}[h]
\centering
\includegraphics[width=0.95\linewidth]{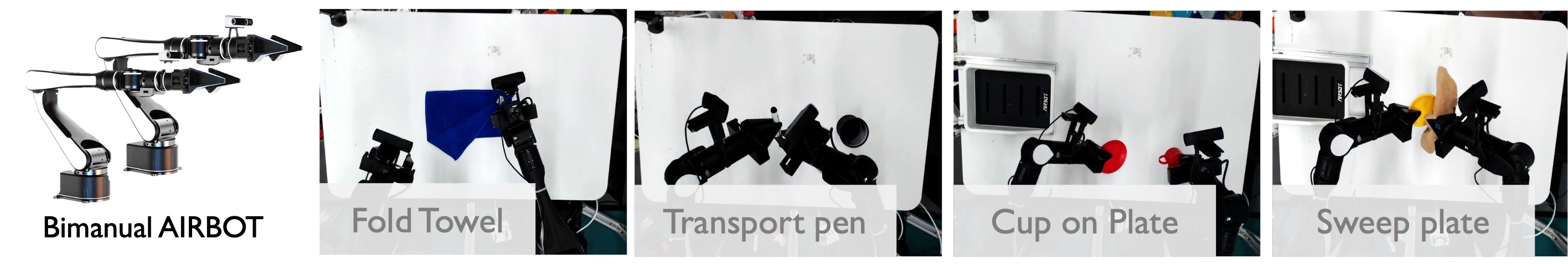}
\end{figure}
\vspace{-0.7cm}
\begin{table}[h]
    \footnotesize
    \setlength{\tabcolsep}{2pt}
    \centering
    \begin{tabular}{c|cccccc}
        Pretrained model& Sweep Plate & Fold Towel & Cup on Plate & Transport pen & \\
       \hline
        LLaVa-OV-0.5B & 7.5& 20 & 2.5 &15 \\
        UniAct-0.5B & \textbf{45} & \textbf{62.5} & \textbf{50} & \textbf{65}
    \end{tabular}
    \vspace{-4pt}
    \caption{Fast adaptation to Bimanual AIRBOT and ACT head}
    \vspace{-10pt}
    \label{tab:Bimanual}
\end{table}

\noindent We further test whether a more advanced decoder can improve performance by integrating ACT~\cite{zhao2023learning} into UniAct-0.5B. We freeze UniAct-0.5B and adapt it to a Bimanual AIRBOT robot, using observations from a top view, two wrist views, and proprioception. We collect 250 demos per task for 4 tasks and compare UniAct-0.5B with the pretrained LLaVa-OV-0.5B model. Tab~\ref{tab:Bimanual} shows UniAct's superior performance, validating its pretraining effectiveness. Additionally, we finetune UniAct-0.5B with the ACT decoder on the LIBERO benchmark, using a top view, wrist view, and proprioception. As shown in Figure~\ref{fig:libero_results}, UniAct achieves further improvements, demonstrating its adaptability with advanced decoders and richer observations. Please refer to Appendix~\ref{sec:fast-adapt-to-bimanual} for details.

\begin{figure*}[t]
    \centering
    \includegraphics[width=0.95\linewidth, height=6.cm]{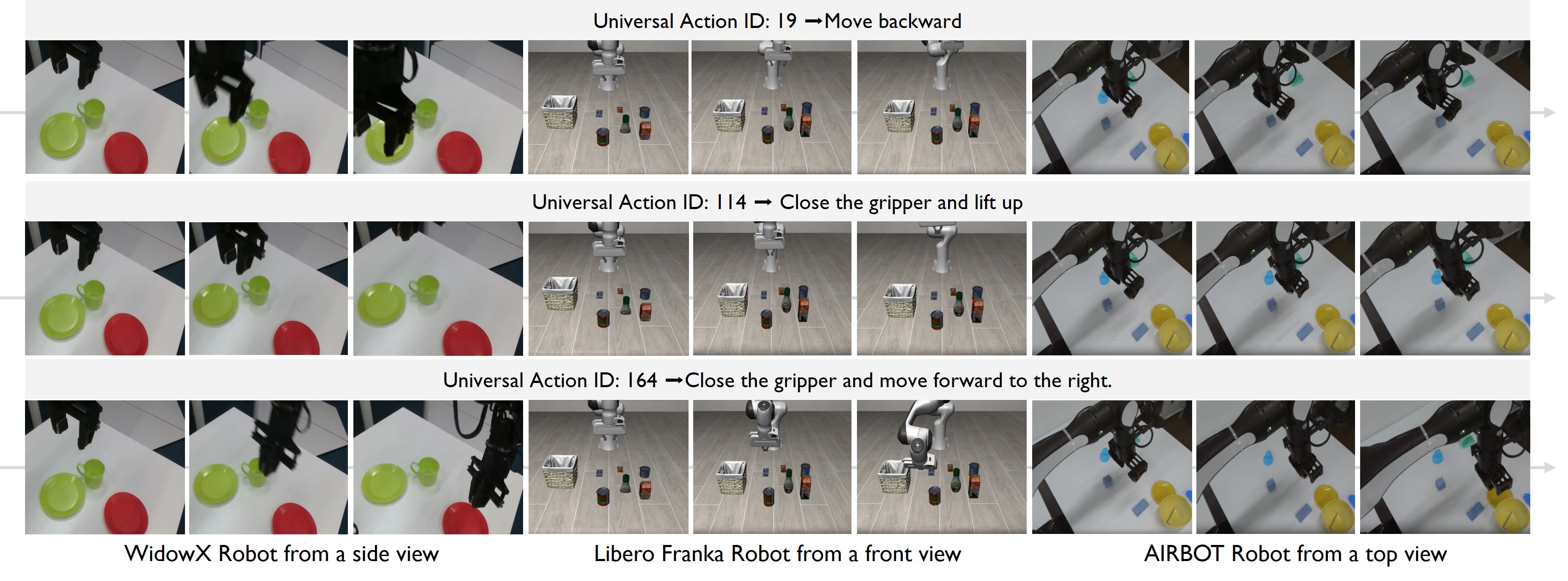}
    \vspace{-13pt}
    \caption{\small We manually check the decoded behavior for all 256 universal actions across different robots and observe at least $40\%$ show exact consistency, even with varying viewpoints and significant domain gaps between the real world and simulation. Visualizations of repeated executions of the same universal action across different robots reveal clear, semantically meaningful, and consistent behaviors.}
    \label{fig:manual_id}
    \centering
    \includegraphics[width=0.95\linewidth, height=5.2cm]{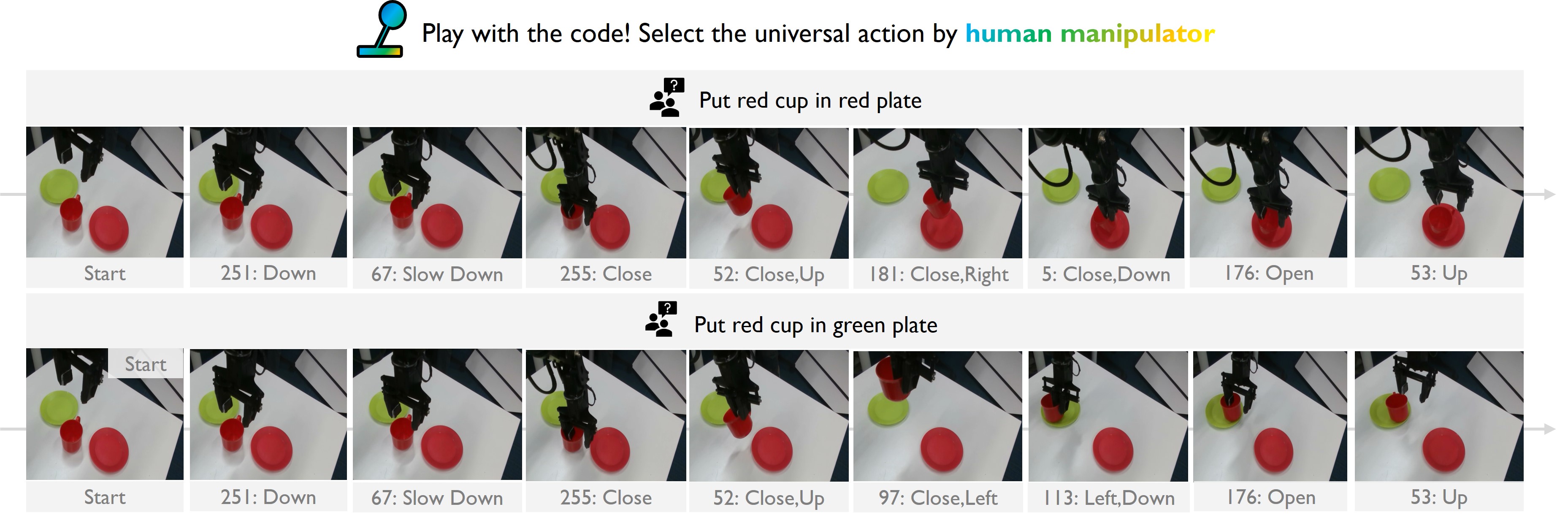}
    \vspace{-7pt}
    \caption{\small Since all universal actions encode semantic meaningful behaviors, we can directly play with it! By selecting the desired behavior using its corresponding universal action ID, we can manually control the robot to perform complex tasks such as \texttt{Pick \& Place}.}
    \vspace{-5pt}
    \label{fig:human_play}
\end{figure*}

\vspace{-1pt}
\subsection{In-depth Analysis of Universal Actions}
\vspace{-1pt}
In this section, we demonstrate that UniAct constructs a meaningful universal action space from two perspectives: 
1) Consistent semantical behaviors are encoded as the same universal actions across diverse embodiments; 2) Universal action extractor can efficiently exploit this shared structure in the universal action space across different robots.

\noindent{\textbf{Interpretation of Universal Actions}}. We manually inspect the decoded behaviors for all 256 universal actions across different robots and observe at least 40\% of them exhibit exact consistency. Fig~\ref{fig:manual_id} shows that the same universal action can be decoded back to consistent behaviors for different robots even with huge gaps. For instance, different robots with different viewpoints, and even those undergoing huge sim-to-real gaps, can execute similar semantical meaningful behaviors given the same universal action.

\noindent{\textbf{Control with the Universal Action}}. Therefore, we can directly interact with the robot to perform desired behaviors by choosing a sequence of universal actions. Fig~\ref{fig:human_play} clearly demonstrates that we can control the robot using universal actions without any robotic knowledge, such as learning complex forward/inverse kinematics transformations. This also underscores the potential of utilizing the universal action extractor as an \textit{action tokenizer} to facilitate future deployments of more advanced embodied foundation models by planning in this discrete universal action space.

\begin{table}[t]
    \centering
    \begin{tabular}{l|cccc}
         & $P_{\rm WidowX}$ & $P_{\rm Franka}$ & $ O_{\rm WidowX}$ & $O_{\rm Franka}$ \\
         \midrule 
        $P_{\rm WidowX}$ & \cellcolor{gray!5} 0.   
                    & \cellcolor{gray!24} {0.34}   
                    & \cellcolor{gray!35} 0.45  
                    &  \cellcolor{gray!51} 0.51  \\
        $P_{\rm Franka}$ & \cellcolor{gray!24} {0.34}
                    &  \cellcolor{gray!5}  0.  
                    &  \cellcolor{gray!60} 0.60  
                    &  \cellcolor{gray!58} 0.58  \\
        $O_{\rm WidowX}$ & \cellcolor{gray!35} 0.45
                    & \cellcolor{gray!60} 0.60
                    & \cellcolor{gray!5} 0.    
                    & \cellcolor{gray!34} {0.44}   \\
        $O_{\rm Franka}$ & \cellcolor{gray!51} 0.51
                    & \cellcolor{gray!58} 0.58
                    & \cellcolor{gray!34} 0.44
                    & \cellcolor{gray!5}0.    \\
    \end{tabular}
    \caption{We calculate the JS divergence of the universal action utilization distributions for two tasks in two distinct domains (lower means more consistent). P and O denote ``pick up the bowl`` and ``open the drawer``, respectively. WidowX and Franka denote the robot in the real world and simulation in Figure~\ref{fig:widowx_results}-\ref{fig:libero_results}, respectively.}
    \label{tab:divergency_in_domain_and_cross_domain}
    \vspace{-0.4cm}
\end{table}

\noindent{\textbf{Statistical Analysis For Universal Action Utilization}}. Here, we summarize the universal action utilization distributions for different tasks across different robots. Table~\ref{tab:divergency_in_domain_and_cross_domain} clearly shows that the utilization distributions for the same tasks and different robots are similar, but for different tasks and same robots are different, demonstrating that the universal action extractor indeed correctly exploits these embodiment-agnostic atomic behaviors by focusing more on task progressions over embodiment details.


\vspace{-1pt}
\section{Conclusion}
\vspace{-1pt}

We introduce UniAct, an innovative embodied foundation modeling framework that operates in a \textit{Universal Action Space} to address the challenge of action heterogeneity. This universal action space encodes shareable atomic behaviors across diverse embodied action spaces to significantly enhance cross-domain data utilization and facilitate cross-embodiment generalization, enabling our 0.5B parameter model to outperform SOTA models that are 14 times larger. Also, the learned universal actions can be precisely translated to any embodiment-specific actions with minimal parameters through heterogeneous decoding, thus allowing for fast adaptation to new robots possessing distinct control interfaces and physical properties. Moreover, our learned universal action extractor can also be used as a universal action tokenizer to power the construction of future large-scale embodied foundation models.
Currently, UniAct is trained with a 0.5B parameter instantiation and evaluated mostly on single-arm robotics platforms due to resource constraints. Future work will focus on scaling up UniAct to larger models and extending its application to a broader range of embodiments, including bi-manual robots and even autonomous driving, further leveraging its versatile capability and effectiveness in more robotic applications.

\section{Acknowledgment}
This work is supported by funding from Beijing Academy of Artificial Intelligence (BAAI) and Wuxi Research Institute of Applied Technologies, Tsinghua University under Grant 20242001120.